\documentclass[11pt]{report}
\newcommand{\method}{SSU\xspace} 
\usepackage{upennstyle} % You may add additional packages below this line. Overleaf will warn you if your packages conflict with the template packages. See Chapter 1 for more information.

\usepackage{amsfonts}
\usepackage{amsmath}
\usepackage{bbm}
\usepackage{subcaption}
\usepackage[nonamebreak,round]{natbib}
\title{Investigating the Feasibility of Mitigating Potential Copyright Infringement via Large Language Model Unlearning}
\author{Guangyao Dou}
\gradgroup{Data Science} % See https://provost.upenn.edu/phd-graduate-groups for names
\date{2024} % Enter current four-digit year
\supervisor{Chris Callison-Burch}
\supervisortitle{Professor of Computer and Information Science}
\gradchair{Susan B. Davidson, Weiss Professor of Computer and Information Science}
\authorlegal{Guangyao Dou} % To opt out of copyrighting your dissertation, uncomment this line. If you do, you will also need to edit line 76.
\cclicense{NonCommercial-ShareAlike 3.0} % If applicable, uncomment this line and edit the text between the {}. If you do, you will also need to edit lines 51, 76, and 78.
\cclicenseurl{https://creativecommons.org/licenses/by-nc-sa/3.0/} % If applicable, uncomment this line and edit the text between the {}. If you do, you will also need to edit lines 49, 76, and 78.
\acknowledgement{
I would like to express my heartfelt gratitude to everyone who contributed to the completion of my master’s thesis and supported me throughout my studies at the University of Pennsylvania.

First and foremost, I am deeply grateful to my thesis advisor, Professor Chris Callison-Burch, for his invaluable guidance in natural language processing and generative models, particularly in the areas of safety and trustworthy AI systems. His Artificial Intelligence course laid a strong foundation in AI and large language models, while his exceptional mentorship has been instrumental in my development as a researcher.

I extend my heartfelt appreciation to the members of Professor Callison-Burch’s lab. I am especially thankful to Veronica Qing Lyu for her insightful suggestions on experimental design and her mentorship, which fostered my scientific rigor and passion for research, ultimately inspiring me to pursue a PhD. I am also deeply grateful to Bryan Li for his constructive feedback on my thesis and to Yikai Mao for his valuable assistance in refining my work.

I am sincerely thankful to Professor Eric Wong for his invaluable feedback on my unlearning research and for his machine learning course, which provided a robust mathematical foundation for exploring the theoretical aspects of my work. I am also grateful to Professor Kaize Ding from Northwestern University for his thoughtful feedback on enhancing both the experimental design and analysis.

My heartfelt gratitude also goes to my dear friend, Zheyuan Liu, who introduced me to the field of machine unlearning, collaborated with me as a coding partner, provided invaluable feedback on my thesis, and shared in the achievement of multiple publications.

Finally, I am profoundly thankful to my parents and sisters for their unwavering support—both financial and emotional—and for encouraging me to pursue my academic interests. Their belief in me has been a constant source of strength throughout this journey.

} % If not applicable, comment out this line to hide the optional acknowledgement page. If you do, you will also need to edit line 86.
\abstract{Pre-trained Large Language Models (LLMs) have demonstrated remarkable capabilities but also pose risks by learning and generating copyrighted material, leading to significant legal and ethical concerns. In a potential real-world scenario, model owners may need to continuously address copyright infringement in order to address requests for content removal that emerge at different time points. One potential way of addressing this is via sequential unlearning, where copyrighted content is removed sequentially as new requests arise. Despite its practical relevance, sequential unlearning in the context of copyright infringement has not been rigorously explored in existing literature. To address this gap, we propose  \textbf{S}table \textbf{S}equential \textbf{U}nlearning (\textbf{SSU}), a novel framework designed to unlearn copyrighted content from LLMs over multiple time steps. Our approach works by identifying and removing specific weight updates in the model’s parameters that correspond to copyrighted content using task vectors. We improve unlearning efficacy by introducing random labeling loss and ensuring the model retains its general-purpose knowledge by adjusting targeted parameters with gradient-based weight saliency. Extensive experimental results show that \method sometimes achieves an effective trade-off between unlearning efficacy and general-purpose language abilities, outperforming existing baselines, but it's not a cure-all for unlearning copyrighted material. \footnote{Code available at \href{https://github.com/guangyaodou/SSU_Unlearn}{guangyaodou/SSU}.}
} % Write your abstract between the {}
\begin{document}
\maketitle % If you are in Romance Languages or Managerial Science and Applied Economics (Wharton), comment out this line. If you do, you will also need to edit lines 33 and 72.
%\makespecializationtitle % If you are in Romance Languages or Managerial Science and Applied Economics (Wharton), uncomment this line. If you do, you will also need to edit lines 33 and 71.
\setcounter{page}{2}

%%%% OPTIONAL COPYRIGHT NOTICE
% \makecopyright % If not applicable, comment out this line to hide the optional traditional copyright notice page. If you do, you will also need to edit line 47.
%-------------------------------------------
\makecreativecommons 
% If applicable, uncomment this line to insert the optional Creative Commons License copyright notice page. If you do, you will also need to edit lines 49, 51, and 76.
%-------------------------------------------

%%%% OPTIONAL DEDICATION PAGE
% \makededication % If not applicable, comment out this line to hide the optional dedication page. If you do, you will also need to edit line 55.
%-------------------------------------------

%%%% OPTIONAL ACKNOWLEDGEMENT PAGE
\makeacknowledgement % If not applicable, comment out this line to hide the optional acknowledgment page. If you do, you will also need to edit line 59.
%-------------------------------------------

\makeabstract
\tableofcontents

%%%% OPTIONAL LIST OF TABLES
\clearpage \phantomsection \addcontentsline{toc}{chapter}{LIST OF TABLES} \begin{singlespacing} \listoftables \end{singlespacing}% If not applicable, comment out this line to hide the optional List of Tables
%-------------------------------------------

%%%% OPTIONAL LIST OF ILLUSTRATIONS
\clearpage \phantomsection \addcontentsline{toc}{chapter}{LIST OF ILLUSTRATIONS} \begin{singlespacing} \listoffigures \end{singlespacing}% If not applicable, comment out this line to hide the optional List of Illustrations
%-------------------------------------------

%%%% OPTIONAL PREFACE
%\makepreface % If applicable, uncomment this line to insert the optional preface. If you do, you will also need to edit line 67.
%-------------------------------------------

%%%%% MAIN TEXT %%%%%
\begin{mainf} 
\chapter{Introduction} % The main body of your dissertation starts below this line
\begin{figure}[t]
    \centering
    \includegraphics[width=0.9\linewidth]{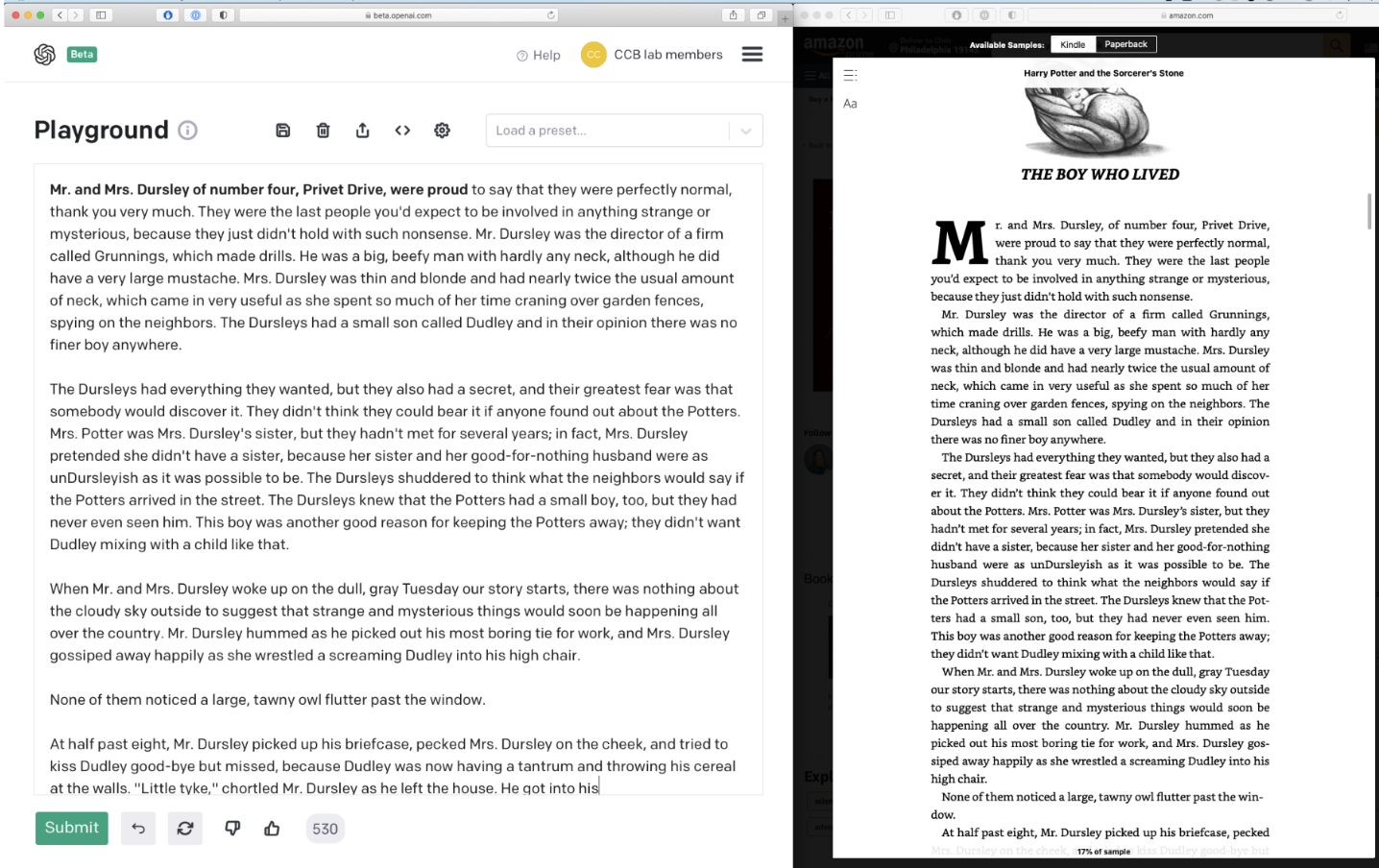}
    % \vspace{-0.2in}
    \caption{An example of a GPT model generating substantially similar copyrighted content from the book Harry Potter, which is highly likely a case of copyright infringement.}
    \label{fig:thumbnail_example}
\end{figure}

In December 2023, the New York Times filed a lawsuit against OpenAI, accusing it of training its Large Language Models (LLMs) on copyrighted material without permission\footnote{\href{https://nytco-assets.nytimes.com/2023/12/NYT_Complaint_Dec2023.pdf}{NYT Complaint, Dec 2023}}. This legal challenge highlights the growing concern over LLMs incorporating copyrighted content from vast pre-training datasets, which are often composed of publicly available text ~\citep{min2023silo, brittain2023us, rahman2023beyond}. Despite the significant progress LLMs have made through learning from diverse text data ~\citep{brown2020language,chowdhery2023palm, touvron2023llama}, screening out copyrighted material remains an immense challenge~\citep{duarte2024cop}. These issues raise broader questions about fair use of generative models. 

There are two times when copyright interacts with LLMs. The first is when LLMs learn from copyrighted materials, which is arguably fair use (but this has not been tested in court). The second is when LLMs generate outputs. If a generated output is substantially similar to copyrighted work it has trained on, then this is more likely to be a copyright infringement. If a court found an AI model developer to be in violation of copyright, then the court may require that the developer to remove that copyrighted work from the model. The cost of retraining from scratch leaving out one copyrighted work is exorbitantly high. Therefore, as an alternate remedy, a court may ask for a copyright takedown that does not require a full retraining of the model. This motivates research into unleanring and other copyright takedown methods. 

Previous works have investigated post-hoc copyright takedown methods -- mitigating risks of generating copyrighted contents -- using system prompt and decoding time intervention such as the MemFree Decoding \citep{ippolito2022preventing}. Additionally, an alternative solution is \textit{machine unlearning}~\citep{cao2015towards}, which removes unwanted knowledge after pre-training, reconfiguring the model as if it had never learned that data. Recent works proposed practical machine unlearning algorithms for LLMs \citep{ yao2023large, eldan2023s, zhang2024negative, chen2023unlearn, jang2023knowledge, zhao2024towards}. However, few have addressed the challenge of \textit{sequentially} unlearning copyrighted contents, leaving it unclear if existing methods are suitable as copyright takedown methods. An effective unlearning algorithm should be \textit{stable}, meaning it should ensure \textit{unlearning efficacy} while preserving non-targeted knowledge, books that are not subject to unlearning, and general-purpose language abilities.

The core of many previous LLM unlearning methods have focused on Gradient Ascent (GA) and further combined it with an in-distribution retained dataset to preserve general-purpose language abilities, known as Gradient Difference~\citep{maini2024tofu, zhao2024towards, liu2024towards, yao2024machine}. However, Gradient Difference requires collection of a substantial amount of in-distribution retained data to maintain general-purpose abilities. Moreover, GA-based methods risk catastrophic collapse, where the model's general-purpose language abilities degrade significantly~\cite{liu2024towards}. \citet{zhang2024negative} proposed Negative Preference Optimization (NPO), framing unlearning as preference optimization. However, NPO relies on a reference model, and if the reference model contains copyrighted information, unlearning efficacy is compromised.

Moreover, the specific challenge of sequential unlearning is that the repeated adjustments to model weights over multiple time steps can disrupt the model, leading to unexpected and sudden degradation. This issue is akin to catastrophic forgetting in continual learning. However, existing unlearning methods struggle in this context because they lack effective control over weight adjustments during the unlearning process, which leads to greater degradation of general-purpose language abilities.

To address these challenges in sequentially unlearning copyrighted books, we propose \textbf{S}table \textbf{S}equential \textbf{U}nlearning (\method), that achieves a better trade-off between effective unlearning and maintaining general-purpose language abilities in sequential settings. Specifically, \method first fine-tunes the model on the copyrighted books, followed by fine-tuning with random labels. During gradient updates, \method applies targeted weight adjustments through gradient-based weight saliency. Afterwards, it extracts task vectors \cite{ilharco2022editing} corresponding to the copyrighted books and subsequently negates these task vectors to achieve unlearning. Unlike Gradient Difference methods, \method does not require additional retained data collection to maintain its performance on other tasks, thereby avoiding the complexity and overhead associated. 

Our experiments on the Llama-3.1-8B-Instruct \citep{dubey2024llama} and Mistral-7B-Instruct-v0.3 \citep{jiang2023mistral} show that \method achieves a superior trade-off between unlearning efficacy and general-purpose language abilities, avoiding the catastrophic collapse observed in GA-based methods. Moreover, \method consistently outperforms NPO, which employs preference optimization frameworks. Additionally, fine-tuning with random loss and targeted model updates play distinct roles in facilitating the unlearning process within \method. In contrast, copyright takedown methods that do not involve model weight updates, such as system prompts and MemFree Decode, fail to effectively mitigate the risks of copyright infringement.

Our main contributions of this thesis are:

\begin{itemize}
   \item We present the first investigation into the sequential unlearning of copyrighted literary works to address copyright infringement, formalizing the task and defining evaluation metrics. 
   \item We systematically evaluate existing methods within the sequential unlearning setting and demonstrate that they either provide limited mitigation of copyright infringement or suffer from catastrophic collapse.
    \item We propose \method, a stable unlearning algorithm for sequential settings, which achieves a superior trade-off between mitigating copyright infringement and preserving reasoning capabilities compared to existing methods.
\end{itemize}

Despite the improved trade-off, unlearning does not fully address copyright takedown requests. Further research is needed to make it an effective remedy.

\chapter{Literature Review}
\section{LLM Memorization and Copyright} 
LLMs are known to sometimes memorize parts of the training data from the pre-training stage~\citep{carlini2021extracting, carlini2022quantifying, zhang2023counterfactual, nasr2023scalable, liu2024shield}. Memorization can result in models outputting copyrighted works from their training data. Although rare, this behavior has prompted recent studies exploring the connection between verbatim memorization and copyright infringement~\citep{chu2024protect, huang2024demystifying, meeus2024copyright, karamolegkou2023copyright}, as well as the fair use of foundation models~\citep{henderson2023foundation}. Various methods have been proposed to mitigate the risk of outputting copyrighted materials verbatim. For instance, \citep{ippolito2022preventing} introduced the MemFree decoding that modifies the generated tokens once identified risks of copyright infringement to reduce the likelihood of copyright infringement but failed to capture non-consecutive verbatim content. Additionally, \cite{min2023silo} proposed SILO, a framework that utilizes a nonparametric datastore containing high-legal-risk data. However, the SILO framework does not address the risks associated with retrieval augmented generation~\citep{wei2024evaluating}, nor does it adequately address the practical challenges of ensuring that the training data for parametric models is limited to permissive content. Additionally, \citet{chiba2024tackling} introduced the PREGen algorithm, which generates prompt variations and selects outputs with minimal originality to further reduce copyright infringement risks. To standardize evaluation, \citet{wei2024evaluating} proposed a comprehensive suite of pipelines for assessing the effectiveness of copyright takedown methods.

\section{Machine Unlearning}
Machine unlearning was first introduced by \citet{cao2015towards}, who proposed using a sharded, isolated, sliced, aggregated (SISA) framework to split the model into smaller sub-models, each learning from a portion of the data. This allows for easier modification of individual sub-models when unlearning is required. There are two main types of unlearning: \textit{Exact Unlearning} and \textit{Approximate Unlearning}. Exact unlearning typically applies to convex settings where all information related to the unwanted data can be completely removed~\citep{ginart2019making, bourtoule2021machine}. In contrast, approximate unlearning is used in non-convex settings and requires the output distribution of the unlearned model to be similar to that of a retrained model from scratch~\citep{guo2020certified, sekhari2021remember, pan2023unlearning, chiencertified}. However, neither exact nor approximate unlearning is applicable to LLMs, as it is infeasible to estimate the output distribution of a LLM. Moreover, \citet{sekhari2021remember} and \citet{zhang2024towards} aimed to connect Differential Privacy \citep{dwork2006differential} and unlearning with theoretical guarantee. However, these are also not applicable to LLMs because many assumptions of DP do not hold in practice. 

Recent advancements in \textbf{Machine Unlearning} have targeted the removal of harmful knowledge~\citep{liu2024towards, yao2023large, li2024wmdp} and copyrighted content~\citep{jang2022knowledge, liu2024breaking, eldan2023s, dou2024avoiding}. Many of these approaches focus on maximizing the loss on datasets designated for forgetting while minimizing the impact on datasets meant to be retained. Building on this foundation, recent studies have introduced unlearning techniques for LLMs through methods such as self-distillation with adjusted logits~\citep{dong2024unmemorization}, data model matching~\citep{georgiev2024attribute}, and entropy maximization~\citep{yuan2024closer}. 

The \textbf{General Data Protection Regulation} (GDPR) of the European Union~\citep{hoofnagle2019european} and the \textbf{California Consumer Privacy Act} (CCPA)~\citep{pardau2018california} have further mandated the \textit{right to be forgotten}~\citep{dang2021right}, encouraging research related to unlearning. The removal of personal information is an additional use case of unlearning beyond copyright. In response, \citet{maini2024tofu} and \citet{yao2024machine} investigated this right, introducing benchmarks to evaluate the effectiveness of private data unlearning for LLMs. Additionally, \citet{liu2024protecting} proposed \textbf{MLLMU-Bench} to examine the performance and challenges of existing methods in unlearning multimodal private datasets, inspiring subsequent work like \textbf{CLIPErase} for removing multimodal data~\citep{yang2024cliperase}.

Despite these efforts, existing studies have not addressed the challenges of unlearning copyrighted works in a sequential setting. Furthermore, the limitations of current unlearning methods in handling sequential unlearning scenarios remain unexplored, highlighting a critical gap in the literature.

\section{Model Editing}

Model editing is a related but distinct process from unlearning~\citep{meng2022mass, hewitt2024model, gupta2024unified}. While unlearning seeks to erase specific information from a model, model editing focuses on modifying or correcting errors without removing the underlying learned knowledge~\citep{liu2024machine}. Recent efforts have aimed to unify existing model editing methods~\citep{gandikota2024unified}, while \citet{gao2024meta} advanced these approaches by integrating them with meta-unlearning frameworks to prevent the relearning of unlearned concepts. 

Additionally, studies have highlighted that model editing in sequential settings can result in model collapse~\citep{gupta2024model, yang2024butterfly}. Model collapse occurs when a model loses its general-purpose language abilities. To address this challenge, potential solutions have been proposed to mitigate the risks associated with sequential edits~\citep{gupta2024rebuilding}.

\chapter{Stable Sequential Unlearning} 
% \label{ch:equations}

\section{Preliminaries}

\subsection{Machine Unlearning for LLMs}

Consider the original model and its weights, denoted as $\theta_o$. Machine unlearning involves the problem where, given a dataset $D = \{(x_i, y_i)\}_{i=1}^N$ that $\theta_o$ was trained on, we aim to intentionally forget a subset of data, denoted as $D_f$, to obtain a modified model, denoted as $\theta_{u}$, such that $\theta_{u}$ behaves as if it has never seen $D_f$ during pre-training. 

In the context of machine unlearning, we often use a retrained model excluding $D_f$ during pre-training as a gold baseline. However, retraining a model for LLMs is extremely expensive and impractical in real-world settings.

\subsection{Task Arithmetic}

Unlearning via negating \textit{task vectors} has recently gained attention~\citep{ilharco2022editing, liu2024towards} and has become an important baseline approach for many unlearning tasks. The rationale behind this approach is that by negating the gradient updates of the unwanted data, we can achieve a more localized unlearning algorithm to effectively erase $D_f$ from $\theta_o$.

Specifically, the process involves two stages. First, we perform standard gradient descent to fine-tune $\theta_o$ on $D_f$, resulting in $\theta_{ft}$. Next, we calculate the task vector as the element-wise difference $\theta_{ft} - \theta_o$. We then negate this task vector from $\theta_o$ to derive the unlearned model $\theta_{u}$, expressed as $\theta_{u} = \theta_o - (\theta_{ft} - \theta_o)$.

\subsection{Unlearning with Multiple Time Steps}

We formally define sequential unlearning as the process where a model, originally trained on a dataset $D$, is incrementally modified to forget subsets of data at multiple time steps while preserving knowledge from the remaining data. Let $D$ be the original dataset, and let $D_f^t \subseteq D$ denote the subset of data that must be forgotten at time step $t$, where $t = 1, 2, \dots, T$. The cumulative set of all data to be forgotten over time is defined as: $D_f = \cup_{t=1}^T D_f^t$. Let $D_r$ represent the subset of data to be retained, such that
$D_r = D \setminus D_f$,  $\ D_f \cap D_r = \emptyset$, and $D_f \cup D_r = D$.

At each time step $t$, the unlearning process modifies the model to forget the subset $D_f^t$, resulting in a sequence of models $\{\theta_u^1, \theta_u^2, \ldots, \theta_u^T\}$. Each $\theta_u^t$ is the model obtained after unlearning the data subsets $D_f^1, D_f^2, \dots, D_f^t$ sequentially. Formally, the goal of sequential unlearning is to ensure that after each unlearning step, the model $\theta_u^t$ minimizes the influence of $D_f^t$ while retaining as much general-purpose knowledge from $D_r$ as possible.

\begin{figure*}[t!]
  \centering 
  \includegraphics[width=1
  \textwidth]{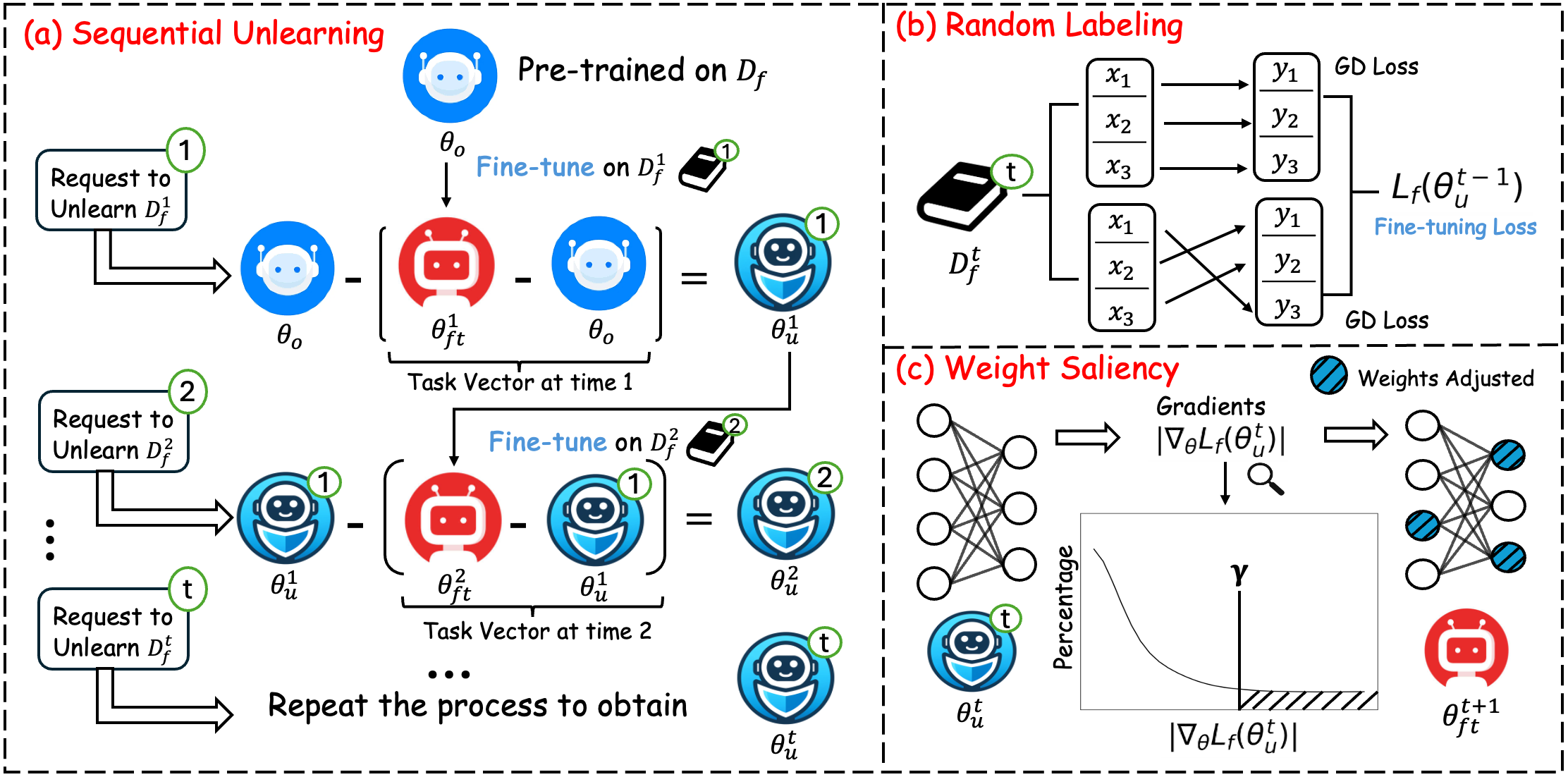}
  \caption{
  \textbf{Overall process of our unlearning framework.} \textbf{(a)} At each time step \( t \), an unlearning request is received to forget the dataset \( D_f^t \). The unlearning algorithm involves first fine-tuning $\theta_{u}^{t-1}$ on $D_f^t$ to obtain $\theta_{ft}^{t}$, and then subtracting the task vector from previously unlearned model $\theta_{u}^{t-1}$. \textbf{(b)} At each time step t. we compute the gradient loss and random labeling loss to obtain the objective $L_f(\theta_{u}^{t-1})$ that will be used for fine-tuning. \textbf{(c)} At time step \(t+1\), we fine-tune $\theta_{u}^{t}$ using the objective we obtained in (b), and only update model weights that are most salient using weight saliency mapping.
  } 
  \label{fig:method_pipeline}
\end{figure*}

\section{Methods}

This section presents \method, which leverages task vectors, incorporates additional loss term for ensuring unlearning efficacy and uses a gradient-based weight saliency map to ensure general-purpose language abilities. The overall process is shown in Figure \ref{fig:method_pipeline}. We first explain unlearning at each time step in section \ref{methods:learning_with_tv}, and then generalize it to sequential setting in section \ref{methods:learning_sequential}.

\subsection{Learning Stable Task Vectors}
\label{methods:learning_with_tv}
First, we present the case of unlearning during the first time step. This means that $t = 1$, $D_f^1 = D_f$, and $\theta_u^0 = \theta_o$. Following the intuition from task vectors, we first need to fine-tune a model on $D_f$. To do this, we define $h_\theta(x, y_{y<i}) = \mathbb{P}(y_i | (x, y_{<i}); \theta)$, which is the probability of the token $y_i$ conditioned on the prompt $x$ and the already generated tokens $y_{<i} = [y_1, y_2, ..., y_{i-1}]$. Next, we define the LLM's loss on $y$ as: \begin{equation}
    L(x, y; \theta) := \sum_{i=1}^{|y|} \ell (h_\theta(x, y_{<i}), y_i),
\end{equation}
in which $l$ is the cross-entropy loss. 

Suppose $\theta_u^0$ is the current LLM through unlearning process. The first goal is to obtain a model that forgets $D_f$. Specifically, we define our first gradient descent loss term as:
\begin{equation}
    \mathcal{L}_{\text{fgt}} = \sum_{(x_{\text{fgt}}, y_{\text{fgt}}) \in D_f} L(x_{\text{fgt}}, y_{\text{fgt}}, \theta_u^0).
\end{equation}

\noindent\textbf{Random Labeling Loss.} Inspired by previous works demonstrating that injecting noise during training improves learning outcomes \citep{miyato2016adversarial, srivastava2014dropout, neelakantan2015adding}, we propose enhancing the effectiveness of unlearning by introducing data augmentation. Specifically, as shown in Figure \ref{fig:method_pipeline} (b), we randomly mismatch the outputs of $D_f$ with the inputs of $D_f$. During the first stage of the task vector approach, we include the following loss:

\begin{equation}
    \mathcal{L}_{\text{rnd}} := \sum_{(x_{\text{fgt}}, ) \in D_{f}} \frac{1}{|D_{f}|} \sum_{(,y_{\text{rnd})} \in D_{f}} L(x_{\text{fgt}}, y_{\text{rnd}}, \theta_t),
\end{equation}
in which $y_{\text{rnd}}$ is any output from $D_f$ and not corresponds to $x_{\text{fgt}}$.

By incorporating this random labeling loss, we introduce controlled noise into the unlearning process. This helps to prevent ``overfitting'' and enhance the stability of unlearning. Combining two loss terms, the final objective can be expressed as: 
\begin{equation}
\label{equation:ssu_main_no_weight_saliency}
    L_f(\theta_t) = \epsilon_1   \mathcal{L}_{\text{fgt}} + \epsilon_2  \mathcal{L}_{\text{rnd}}.
\end{equation} 

\noindent\textbf{Weight Saliency.} One of the main challenges with existing methods is that they often fail to control the changes in model weights during this process, which can lead to instability and faster degradation of the model's general-purpose capabilities.

To preserve general-purpose language abilities, it is essential to mitigate the risk of catastrophic collapse during each time step of sequential unlearning. We can achieve this by steering the unlearning process towards specific parts of the model weights that are most relevant to the data to be forgotten. 
As shown in Figure \ref{fig:method_pipeline} (c), we use a weight saliency map \citep{fan2023salun} during the first stage of fine-tuning to further ensure localized unlearning by only adjusting specific weights that are most influenced by the data to be forgotten. The weight saliency map is defined as:
\begin{equation}
\label{equation:ssu_main_weight_saliency_mask}
     m_s = \mathbbm{1}(\lvert \nabla_{\theta} L_f (\theta_t) \rvert \geq \gamma),
\end{equation}
in which $\mathbbm{1}(f \geq \gamma)$ is an element-wise indicator function which outputs one for the $i$-th element if $f_i \geq \gamma$, and 0 otherwise, and $\nabla_{\theta} L_f (\theta_t)$ is a gradient vector. 

Next, we apply the weight saliency mapping on the parameters that are most salient to unlearning and have the learned model as at each gradient accumulation step as:
\begin{equation}
\label{equation:objective_equation}
    \theta_{t+1} = m_s \odot (\Delta\theta + \theta_t) + (1 - m_s) \odot \theta_t,
\end{equation}
where $\Delta\theta$ indicates model updates. After training for $T$ gradient accumulation steps using Equation \ref{equation:objective_equation}, we obtain a fine-tuned model $\theta_{ft}^1$. Finally, we obtain our modified model using task vector by negating the knowledge of $D_f$ learned during the fine-tuning process from the original model as:
\begin{equation}
    \theta_u^1 = \theta_o - (\theta_{ft}^1- \theta_{o}).
\end{equation}

\subsection{Sequential Unlearning}
\label{methods:learning_sequential}
In this section, we explain how we generalize unlearning to sequential setting. As shown in Figure \ref{fig:method_pipeline}, at each new time step \( t \), we have the previously unlearned model $\theta_{u}^{t-1}$. Once we receive a new unlearning request, we will fine-tune $\theta_{u}^{t-1}$ using equation \ref{equation:objective_equation} as discussed in section \ref{methods:learning_with_tv} and obtain $\theta_{ft}^{t}$. Lastly, we obtain a new unlearned model at time step \( t \) as:
\begin{equation}
    \theta_{u}^t = \theta_{u}^{t-1} - (\theta_{ft}^t - \theta_{u}^{t-1}).
\end{equation}
If more unlearning requests are received, we will iteratively apply the same process to obtain a newer unlearned model. Notably, unlike Gradient Difference, \method does not require any additional retained dataset when calculating equation \ref{equation:ssu_main_no_weight_saliency}, ensuring efficiency and simplicity in real-world applications.

\chapter{Experiment and Evaluation} 
\section{Experimental Setup}
In this thesis, we choose the removal of copyrighted books from LLMs as a representative scenario for authors to exercise their \textit{intellectual property rights}. Machine unlearning can be applied to these LLMs to unlearn these books, thereby preventing the generation of verbatim copyrighted content.

\subsection{Setting}
We aim to evaluate the effectiveness of sequential unlearning of copyrighted books. At each time step, we unlearn one book following the experimental design of \citep{zhou2023making, carlini2022quantifying}. For each book, we split the text into chunks of 200 tokens and use the system prompt, instruction prompt, and the first 100 tokens as prompt text to ask the model to continue the story, with the following 100 tokens serving as the ground truth. To assess the amount of copyrighted information potentially being reproduced, we compared the LLM's completion with the remaining 100 tokens of each chunk from the original book using techniques for extract training data proposed by \cite{yu2023bag}. 

To evaluate the effectiveness of sequential unlearning, we conduct experiments on several copyrighted books. Our process involves the following steps:

First, each book is split into many chunks of 200 tokens. For each chunk, the initial 100 tokens are used as a prompt, which is fed into the LLM. The remaining 100 tokens serve as the answer or continuation from the original book. This setup allows us to assess how well the model can generate text that follows the given prompt.

In addition to the prompt from the book, we use a system prompt and a instruction prompt to guide the model in generating the completion. 

Following \cite{wei2024evaluating}, the default system prompt for all of our experiments is \footnote{This prompt was designed to make chatbots more responsive and effective in following instructions.} 

\begin{quote} "You are a helpful, respectful and honest assistant." \end{quote}

and the default instruction prompt is  

\begin{quote} "Please complete the rest of the following paragraph based on the context." \end{quote}

For each prompt, the model generates a completion using a nucleus sampling by setting the temperature to 0.4 and $\eta = 0.6$. This follows bags of tricks to extract training data suggested by \citet{yu2023bag}.

To evaluate the generated completions and its risk of copyright infringement, we use Rouge-1 and ROUGE-L score. These metrics allow us to compare the LLM's completions with the original text and assess the model's ability to unlearn specific content.

Specifically, we evaluate the scores on the following sets of books:
\begin{itemize}
\item Books to be forgotten ($D_f$)
\item Books that are previously unlearned when time step is greater than one ($D_{prev}$)
\item Books that not to be forgotten ($D_{nor}$) 
\end{itemize}

\subsection{Evaluation Metrics}
In the United States, the fair use of copyrighted work does not constitute a violation of copyright law \citep{karamolegkou2023copyright}. A key aspect of the fair use defense is the degree of transformativeness and the amount of the original work used \citep{asay2020transformative}. While a de minimis amount of copying is permissible \citep{henderson2023foundation}, we aim to ensure that the outputs generated by LLMs do not replicate a substantial portion of any copyrighted works. To minimize the risk of copyright infringement, we assess how substantially similar the generated outputs are in comparison to the original continuations that are present in the copyrighted work.

Following \citet{wei2024evaluating} and \citet{shi2024muse}, we use Rouge-1 and Rouge-L scores \cite{lin2004rouge} to measure the similarities between the model's outputs and the original content. For the books we aim to unlearn, lower Rouge-1 and Rouge-L scores indicate greater   transformativeness, thereby reducing the risk of copyright infringement. In our experiments, we evaluated Rouge-1 and Rouge-L scores on the datasets $D_f$, $D_{prev}$, and $D_{nor}$. \footnote{In alignment with previous copyright evaluation metrics, and recognizing that semantic similarity alone is insufficient to determine copyright infringement, we include evaluation metrics that focus on lexical similarity (e.g., Rouge) and exclude those that solely reflect semantic similarity.}

In addition to evaluating the model's unlearning effectiveness, we also assessed general-purpose language abilities after copyright takedown methods. The tasks considered include Massive Multitask Language Understanding (MMLU) \citep{hendrycks2020measuring} and MT-Bench \citep{zheng2023judging}. 

\subsubsection{Rouge-1} Recall-Oriented Understudy for Gisting Evaluation (Rouge) includes Rouge-1, which measures the overlap of unigram (single word) occurrences between the LLM's completion and the original books. A unigram is any individual word that appears in both the completion (hypothesis text) and the original book (reference text).

First, we define the recall as the ratio of the number of overlapping unigrams between the hypothesis and reference text to the total number of unigrams in the reference text: \begin{equation} Recall = \frac{\text{overlapping unigrams}}{\text{total unigrams in the reference text}}. \end{equation}

Similarly, precision is defined as the ratio of the number of overlapping unigrams to the total number of unigrams in the hypothesis text: \begin{equation} Precision = \frac{\text{overlapping unigrams}}{\text{total unigrams in the hypothesis text}}. \end{equation}

Lastly, the Rouge-1 score used in our experiments is calculated as the F1 score, which combines precision and recall: \begin{equation} F1 = 2 \cdot \frac{Precision \cdot Recall}{Precision + Recall} \end{equation}

\subsubsection{Rouge-L}
Rouge-L measures the longest common subsequence (LCS) between the LLM's completion and original books. In detail, LCS is a sequence that appears in both the completion (hypothesis text) and original book (reference text) in the same order but not necessarily contiguously. 

Next, we define the recall as the ratio of the length of the LCS to the total length of the reference text:
\begin{equation}
Recall = \frac{LCS}{\text{length of the reference text}}.
\end{equation}

We also define the precision as the ratio of the length of the LCS to the total length of the hypothesis text:
\begin{equation}
Precision = \frac{LCS}{\text{length of the hypothesis text}}.
\end{equation}

Lastly, the Rouge-L score we used in our experiments is defined as:
\begin{equation}
F1 = 2 \cdot \frac{Precision \cdot Recall}{Precision + Recall}
\end{equation}

\subsection{Datasets and Models} We use the open-source language models Llama-3.1-8B-Instruct (Llama3.1) \citep{dubey2024llama} and Mistral-7B-Instruct-v0.3 (Mistral-7B) \citep{jiang2023mistral}, both fine-tuned on a dataset of 10 books from Project Gutenberg \footnote{\href{https://www.gutenberg.org/}{gutenberg.org}} ($D_f$) for one epoch as the vanilla models for our experiments.

For Llama-3.1, we unlearned 10 books across 10 time steps, while for Mistral-7B, we unlearned only the first six books, as many methods collapsed by time step 6. As a proxy for copyrighted books, we selected several public domain texts, including \textit{The Adventures of Sherlock Holmes} by Arthur Conan Doyle, \textit{Pride and Prejudice} by Jane Austen, and \textit{Alice’s Adventures in Wonderland} by Lewis Carroll, along with seven other popular books. These books are listed on Project Gutenberg, making it likely that they were included in the model’s training data. Additionally, we selected books to represent knowledge we aim to preserve after unlearning. These books were randomly sampled from Project Gutenberg and are denoted as $D_{nor}$. All books were pre-processed following the methodology of \citet{gerlach2020standardized}. For $t > 1$, we constructed $D_{prev}$ by aggregating all previously unlearned books.

\subsubsection{Books to Forget}
We crawled all available books from Project Gutenberg and pre-processed them following the methodology of \citet{gerlach2020standardized}, where we remove all headers and boiler plate text.

At time step 1, we unlearn The Adventures of Sherlock Holmes by Arthur Conan Doyle. At time step 2, we unlearn Flowers of the Sky by Richard A. Proctor, Pagan Papers by Kenneth Grahame at time step 3, Diary of Samuel Pepys by Samuel Pepys at time step 4, Pride and Prejudice by Jane Austen at time step 5, They Call Me Carpenter: A Tale of the Second Coming by Upton Sinclair at time step 6, Memoirs of the Court of Louis XIV. and of the Regency — Complete by Orléans at time step 7, Alice's Adventures in Wonderland by Lewis Carroll at time step 8, The Wonderful Adventures of Nils by Selma Lagerlöf at time step 9, and Starr, of the Desert by B. M. Bower at time step 10. 

For the books in $D_f$, the entire texts are split into chunks of 200 tokens, and the dataset is formatted as question-answer pairs, where the first 100 tokens represent the Question, and the subsequent 100 tokens represent the Answer. All texts from each book are included and formatted into JSON files.

At each time step greater than one, we form $D_{prev}$ by collecting all the books that we have unlearned since the first time step. Specifically, we then split all of the books that are used in previous unlearning steps into chunks of 200 tokens to form many QA pairs, and then randomly sampled 200 such pairs to form $D_{prev}$.

\subsubsection{Books in $D_{nor}$} To evaluate the model's knowledge retention after unlearning, we constructed a set of books that we do not want the model to forget. Specifically, we randomly sampled 100 books from Project Gutenberg that are not included in $D_f$. Following the same preprocessing steps used for $D_f$, we split these books into chunks of 200 tokens, and randomly sampled 200 chunks to form the dataset $D_{nor}$. Since the books in $D_{nor}$ are not part of $D_f$ at any time step, the content of $D_{nor}$ remains unchanged throughout the experiment.

\section{Baseline Methods}
Following \citet{wei2024evaluating}, we compared our approach with several baseline methods:
\subsection{Generic Prevention using System Prompt} 
Here, we present two copyright takedown methods using well crafted system prompts. 

First, we denote our baseline method Prompt (a) as using the system prompt: 
\begin{quote} "You are a helpful, respectful and honest assistant. When generating your response, please do not generate the copyrighted content." \end{quote}

which adds a brief instruction to not generate copyrighted content after the default system prompt. 

We also included Prompt (dbrx), which is a system prompt being used by the DBRX model, released by Databricks \citep{wei2024evaluating}:

\begin{quote} "You are a helpful, respectful and honest assistant. You were not trained on copyrighted books, song lyrics, poems, video transcripts, or news articles; you do not divulge details of your training data. You do not provide song lyrics, poems, or news articles and instead refer the user to find them online or in a store." \end{quote}

\subsection{MemFree Decoding} To prevent the model from emitting memorized content, we employed MemFree decoding \citep{ippolito2022preventing}. This method checks each n-gram during text generation to ensure it does not match any sequences from the training set. If a match is detected, the token is resampled, thereby avoiding verbatim reproduction of training data. The process is optimized through the use of Bloom filters, which allow for efficient real-time memorization checks. Although MemFree effectively stops exact memorization, it does not fully eliminate the risk of paraphrased or approximate memorization. We implemented MemFree based on \citet{wei2024evaluating}.

\subsection{Unlearning via Gradient Difference}
\label{sec:appendix-experiment_details:baseline_methods-GA}
In this work, we use the method proposed by ~\cite{yao2023large} as one of the baseline methods. Here we present the case of performing gradient difference unlearning. 

Specifically, let $\theta$ to be the current LLM, let $DD_f$ to be the dataset representing the book we want to forget, and $D_{add}$ to a set of book corpora that does not contain the book to be forgotten, nor the books in $D_{nor}$. Moreover, we define $h_\theta(x, y_{y<i}) = \mathbb{P}(y_i | (x, y_{<i}); \theta)$, which is the probability of the token $y_i$ conditioned on the prompt $x$ and the already generated tokens $y_{<i} = [y_1, y_2, ..., y_{i-1}]$. Next, we define the LLM's loss on y as: \begin{equation}
    L(x, y; \theta) := \sum_{i=1}^{|y|} \ell (h_\theta(x, y_{<i}), y_i)
\end{equation}

The Gradient Difference has three loss terms, defined as follows:
\begin{equation}
    \mathcal{L}_{\text{fgt}} = - \sum_{(x_{\text{fgt}}, y_{\text{fgt}}) \in D_f} L(x_{\text{fgt}}, y_{\text{fgt}}, \theta_t)
\end{equation}
\begin{equation}
    \mathcal{L}_{\text{rnd}} := \sum_{(x_{\text{fgt}}, ) \in D_{f}} \frac{1}{|Y_{\text{rnd}}|} \sum_{(,y_{\text{rnd})} \in Y_{\text{rnd}}} L(x_{\text{fgt}}, y_{\text{rnd}}, \theta_t)
\end{equation}
\begin{equation}
    \phi_\theta = h_{\theta} (x_{\text{nor}}, y_{\text{nor}<i})
\end{equation}
\begin{equation}
    \mathcal{L}_{\text{add}} := \sum_{(x_{\text{add}}, y_{\text{add}}) \in D_{\text{add}}} \sum_{i=1}^{|y_{\text{add}}|} \text{KL}(\phi_{\theta_o} \parallel \phi_{\theta_t}).
\end{equation}
in which $Y_{\text{rnd}}$ is a set of responses irrelevant to responses of $D_f$, sampled from $D_{add}$. 

Lastly, the GA approach is trying to minimize the following loss to obtain the unlearned model:
\begin{equation}
    L = \epsilon_1 \mathcal{L}_{\text{fgt}} + \epsilon_2  \mathcal{L}_{\text{rnd}} + \epsilon_3  \mathcal{L}_{\text{add}}
\end{equation}
\begin{equation*}
    \theta_{t + 1} \leftarrow \theta_t - \nabla L.
\end{equation*}
in which $\mathcal{L}_{\text{fgt}}$ is a gradient ascent loss on $D_f$, which tries to make the model perform poorly on the $D_f$. Next, $\mathcal{L}_{\text{rnd}}$ tries to randomly mismatch the labels from non-relevant dataset to the inputs of the dataset we want to forget. Lastly, $\mathcal{L}_{\text{add}}$ tries to maintain the performance on the normal dataset. In our experiment, we set $\epsilon_1 = 1$, $\epsilon_2 = \epsilon_3 = 0.5$ across all time steps and all models.

\subsection{Unlearning via Task Vector}
We also use the task vector method as one of the baseline approaches, which typically involves a two-stage process. Considering the case of $t = 1$, we denote $\theta_o$ as the original model weights. We intentionally fine-tune the model on $D_f$ to obtain $\theta_{ft}^1$. This fine-tuning process is defined as follows:

\begin{equation}
    \mathcal{L}_{\text{fgt}} = \sum_{(x_{\text{fgt}}, y_{\text{fgt}}) \in D_f} L(x_{\text{fgt}}, y_{\text{fgt}}, \theta_t)
\end{equation}
\begin{equation}
    \theta_{t+1} \leftarrow \theta_t - \epsilon \nabla_{\theta_t} \mathcal{L}_{\text{fgt}}
\end{equation}

Next, we define the task vector $\tau$ as the element-wise difference between $\theta_{ft}$ and $\theta_o$:

\begin{equation}
    \tau = \theta_{ft}^1 - \theta_o
\end{equation}

Finally, the unlearned model $\theta_u$ at time step $t$ is obtained by:

\begin{equation}
    \theta_u^1 = \theta_o - \tau
\end{equation}

The general intuition behind this method is to first obtain a model that is specialized in the dataset we aim to forget. The task vector $\tau$ represents the changes in weights required to acquire this specific knowledge. By subtracting these "knowledge" weights from the original model, we effectively unlearn the targeted information.

\subsection{Unlearning via NPO}
\label{sec:appendix-experiment_details:unlearning_NPO}

In this work, we utilize the Negative Preference Optimization (NPO) method for unlearning undesirable data, aiming to overcome the catastrophic collapse often observed with gradient ascent methods. NPO builds on the framework of preference optimization, specifically focusing on negative samples from the dataset to be unlearned.

The NPO loss function is defined as follows:
\begin{equation}
    \mathcal{L}_{\text{NPO}} = \frac{2}{\beta} \mathbb{E}_{(x, y) \in D_{\text{forget}}} \left[ \log \left(1 + \left(\frac{\pi_\theta(y|x)}{\pi_{\text{ref}}(y|x)}\right)^\beta \right) \right]
\end{equation}
where \( \pi_\theta(y|x) \) represents the prediction probability of the current model for token \( y \) given the input \( x \), and \( \pi_{\text{ref}}(y|x) \) is the prediction probability from the reference model trained on the entire dataset. The parameter \( \beta \) controls the smoothness of the optimization, and as \( \beta \to 0 \), the NPO loss converges to the standard gradient ascent loss. 

Minimizing this loss helps reduce the model’s reliance on the forget set, ensuring that the unlearning process remains stable and avoids the rapid deterioration seen in gradient ascent approaches. In our experiment, we set $\beta = 0.4$, and we obtain $\pi_{\text{ref}}$ by optimizing off-the-shelf LLMs on $D_{f} \bigcup D_{nor}$. 

\subsection{Implementation Details}
The experiments are conducted on eight RTX A6000 GPUs. For all unlearning algorithms, at each time step, we only unlearn the model for 1 epoch, with a batch size set to 2. During all the fine-tuning process, we used Lora \citep{hu2021lora}, and we did not quantize the model because quantization leads to inaccurate element-wise subtraction for TV-based methods. 

For the Llama-3.1-8B-Instruct model, we set the learning rate to be 1e-5 for the first five time steps, and decrease the learning rate to 1e-6 for the rest of the time steps. 

For the Mistral-7B-Instruct-v0.3, we set the learning rate to be 1e-5 for the first three time steps, and the learning rate to be 1e-6 for the rest of time steps. 

\noindent\textbf{SSU.} For SSU, we set $\epsilon_1 = 1$,a and $\epsilon_2 = 0.5$ for all the time steps and models. We set $\gamma$ to be 1 standard deviation away from the  mean of the gradient vector $\nabla_{\theta} L_f (\theta_t)$. 

\section{Results}
We present experimental results of Llama3.1 for different unlearning time steps in Figure \ref{fig:main_book_forget_all_appendix}.  See the results for Mistral-7B in Figure \ref{fig:main_book_forget_all_mistral}. 

\subsection{Sequential Unlearning of Books}

This section examines the impact of unlearning on \( D_{f} \) and $D_{prev}$. Ideally, the model should have lower average Rouge scores to demonstrate lower risks of copyright infringement. 

\paragraph{\method consistently ranks among the top unlearning methods for mitigating copyright infringement.}
As shown in Figure \ref{fig:main_book_forget_unlearn}, for Llama3.1,  \method consistently achieves one of the lowest average Rouge scores on $D_f$. Similarly, as shown in Figure \ref{fig:main_book_forget_previous}, \method proves to be one of the most effective across all time steps for books in $D_{prev}$, more effective in mitigating copyright infringement than NPO and Gradient Difference. The results are similar for Mistral-7B. As shown in Figures \ref{fig:main_book_forget_unlearn_mistral} and \ref{fig:main_book_forget_previous_mistral}, Gradient Ascent and Gradient Difference achieve the lowest Rouge scores on $D_f$ at later time steps, effectively erasing copyrighted content through opposite gradient updates, ultimately leading to catastrophic collapse. TV also maintains effective unlearning, but also collapses at time step 5. In contrast, NPO’s average Rouge score on $D_f$ remains close to the vanilla model, indicating ineffectiveness in mitigating copyright infringement.  On the other hand, \method demonstrates less risks of copyright infringement.

\paragraph{Prompting and MemFree Decoding offer limited mitigation of copyright infringement.} As shown in Figures \ref{fig:main_book_forget_unlearn} and \ref{fig:main_book_forget_unlearn_mistral}, the Rouge scores for prompting and MemFree are largely indistinguishable from the vanilla model across many time steps and models. Although system prompts attempt to prevent generating copyrighted content, Llama3.1 often produces higher Rouge scores with prompting (a), and prompting (dbrx) is only marginally effective at certain time steps. We suspect this is due to the instruction-tuned Llama3.1 model’s inability to differentiate what constitutes copyrighted content. For the Mistral-7B model, as shown in Figures \ref{fig:main_book_forget_unlearn_mistral} and \ref{fig:main_book_forget_previous_mistral}, both prompting methods are similarly ineffective in reducing infringement risks.  Lastly, the MemFree decoding is always ineffective in unlearning copyrighted books for for Llama3.1 and Mistral-7B.

\begin{figure*}
\centering
         \begin{subfigure}[b]{\textwidth}
            \centering
            \includegraphics[width=1\textwidth]{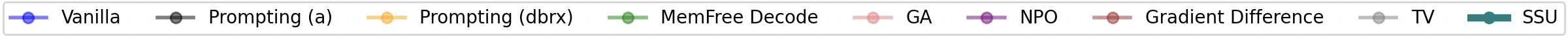}
          \end{subfigure}
        \begin{subfigure}{0.465\textwidth}
        \includegraphics[width=\textwidth]{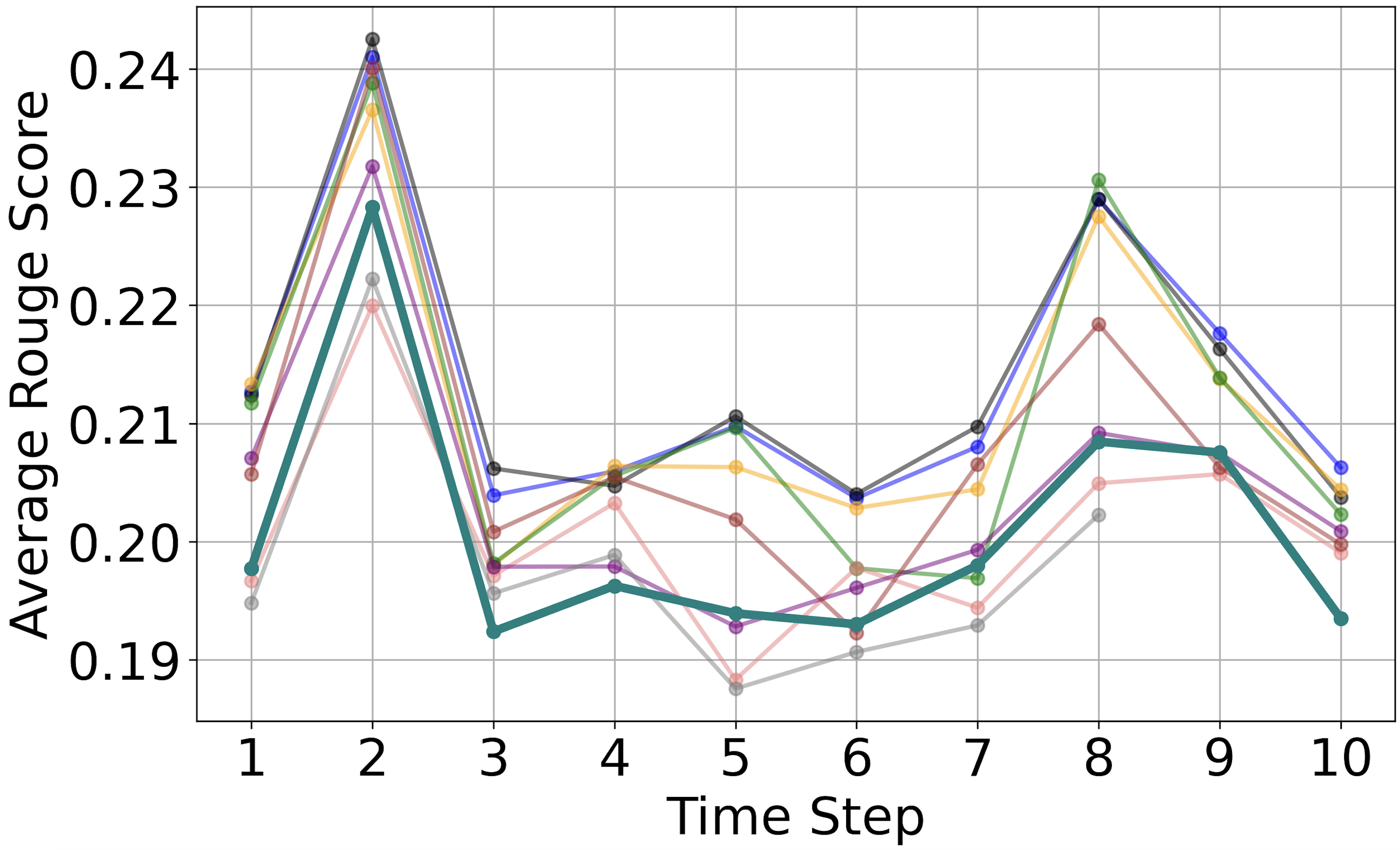}
        \subcaption{Avg Rouge Score on $D_f$}
        \label{fig:main_book_forget_unlearn}
        \end{subfigure}
        \begin{subfigure}{0.465\textwidth}
        \includegraphics[width=\textwidth]{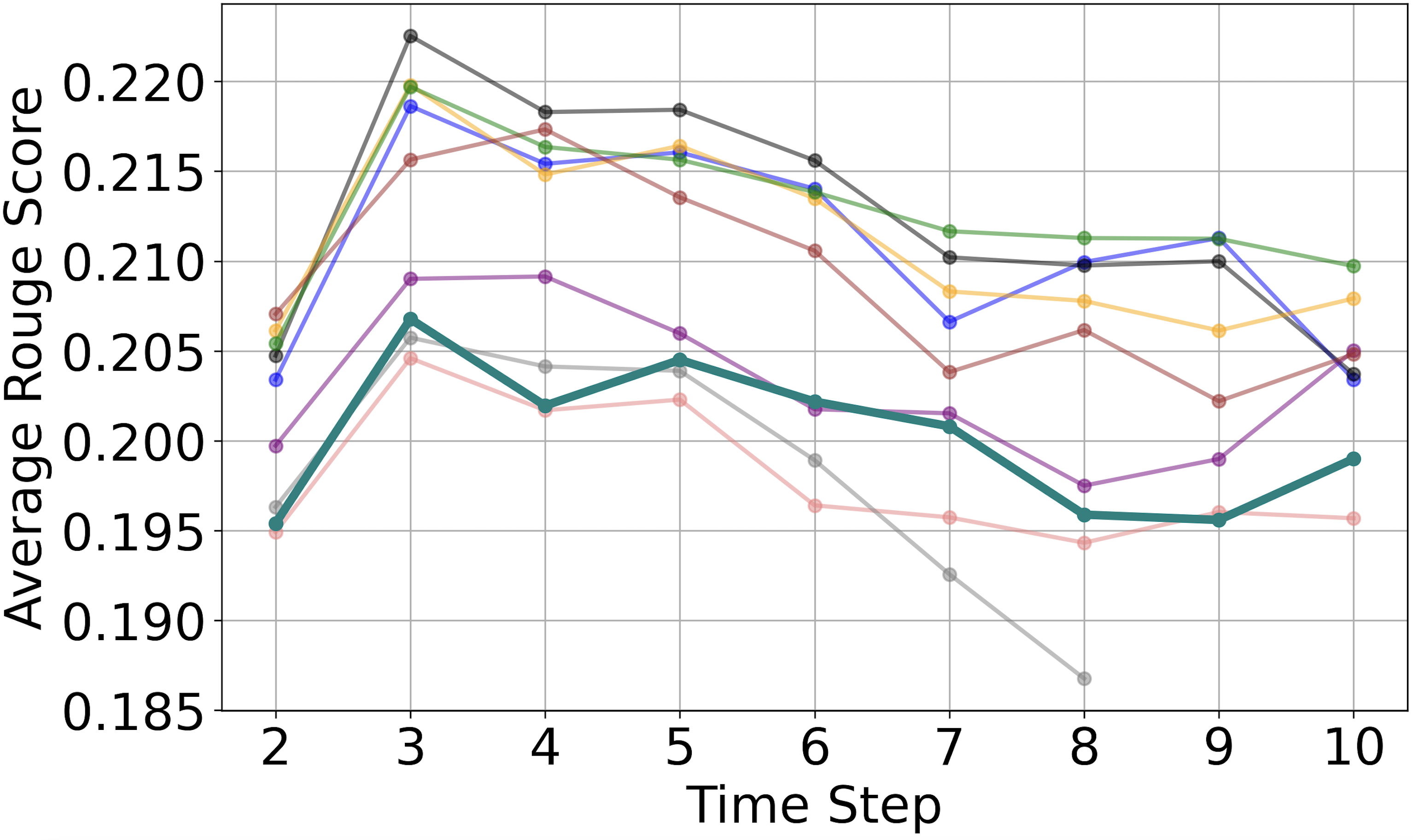}
        \subcaption{Avg Rouge Score on $D_{prev}$}
        \label{fig:main_book_forget_previous}
        \end{subfigure} 
        
      \begin{subfigure}{0.480\textwidth}
        \includegraphics[width=\textwidth]{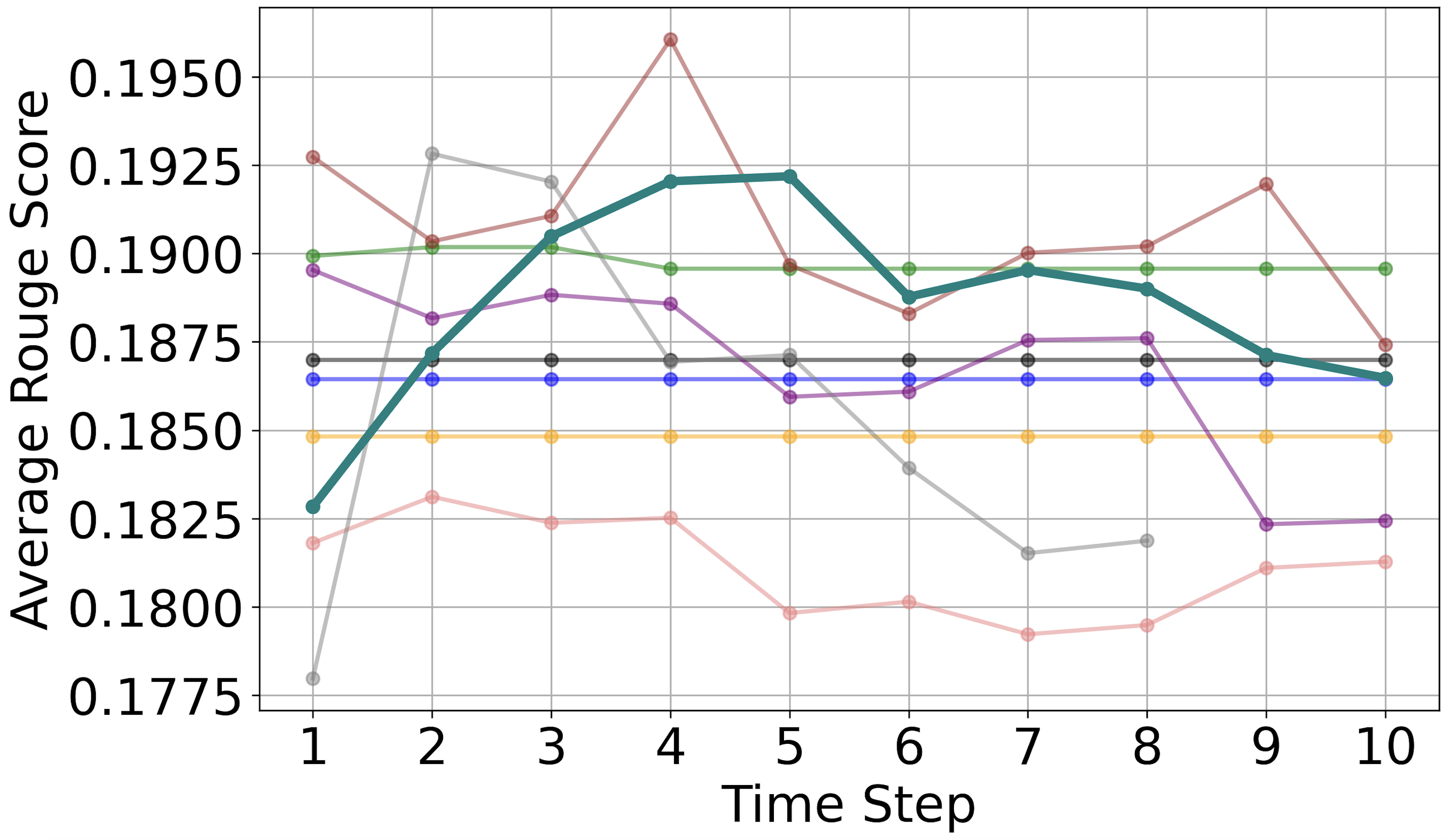}
        \subcaption{Avg Rouge Score on $D_n$}
        \label{fig:main_book_forget_normal}
        \end{subfigure}
        \begin{subfigure}{0.450\textwidth}
        \includegraphics[width=\textwidth]{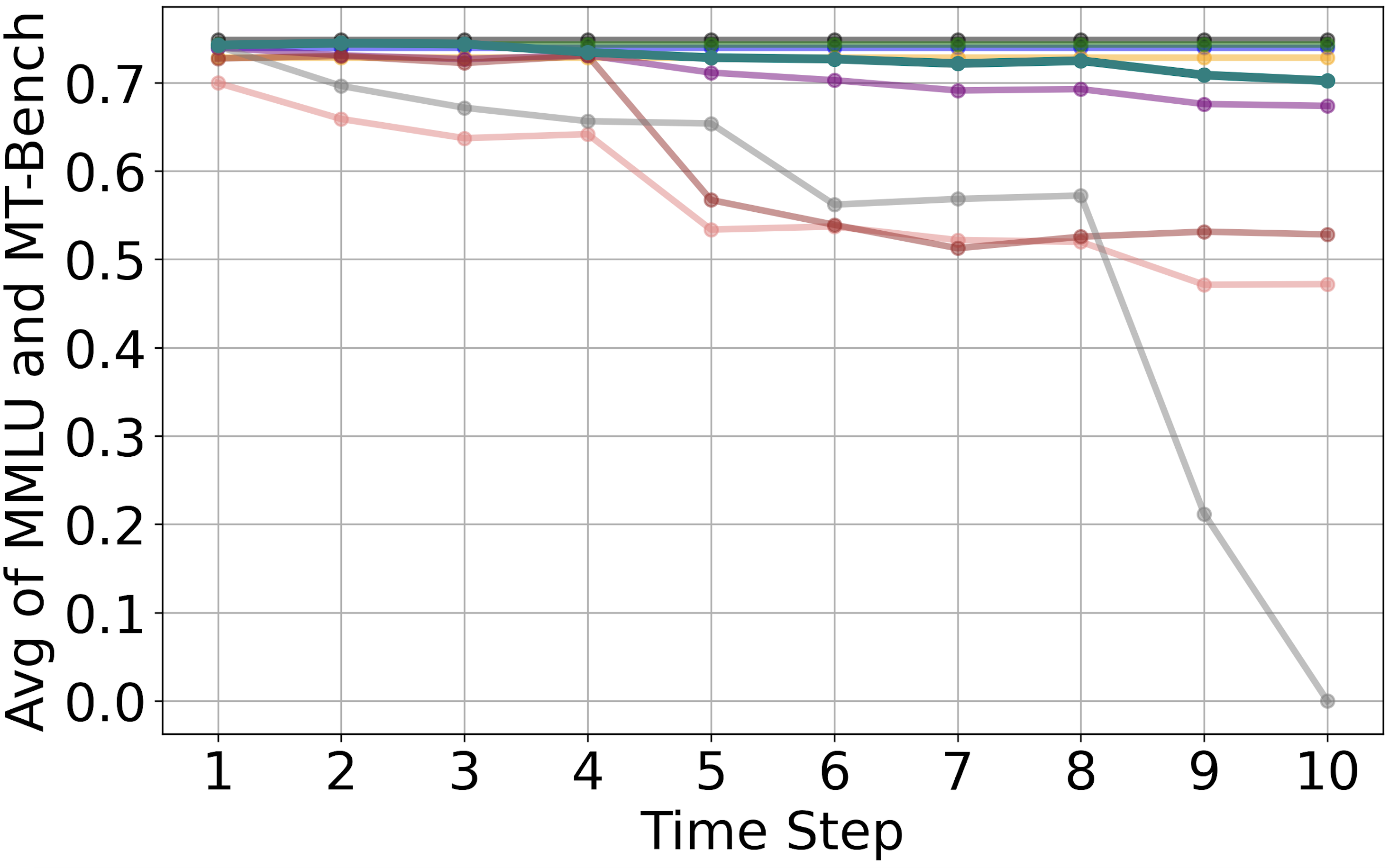}
        \subcaption{Avg MMLU and MT-Bench score}
\label{fig:main_book_forget_general_llama3.1}
        \end{subfigure}
\caption{The average of Rouge-1 and Rouge-l and benchmark scores for LLaMA3.1: (a) books to forget $D_f$ ($\downarrow$); (b) previously unlearned books $D_{prev}$ ($\downarrow$); (c) $D_{nor}$ ($\uparrow$). and (d) averaged normalized MMLU and MT-Bench scores ($\uparrow$). The results for TV after time step 8 are omitted due to collapse. Lower Rouge scores for $D_f$ and $D_{prev}$ indicate better unlearning, while higher scores for $D_{nor}$ and benchmarks reflect better performance.}
\label{fig:main_book_forget_all_appendix}
\end{figure*}

\begin{figure*}
\centering
        \begin{subfigure}[t]{1\textwidth}
            \centering
            \includegraphics[width=1\textwidth]{Figure/save_figure/legend_plot.png}
          \end{subfigure}
        \begin{subfigure}{0.48\textwidth}
        \includegraphics[width=\textwidth]{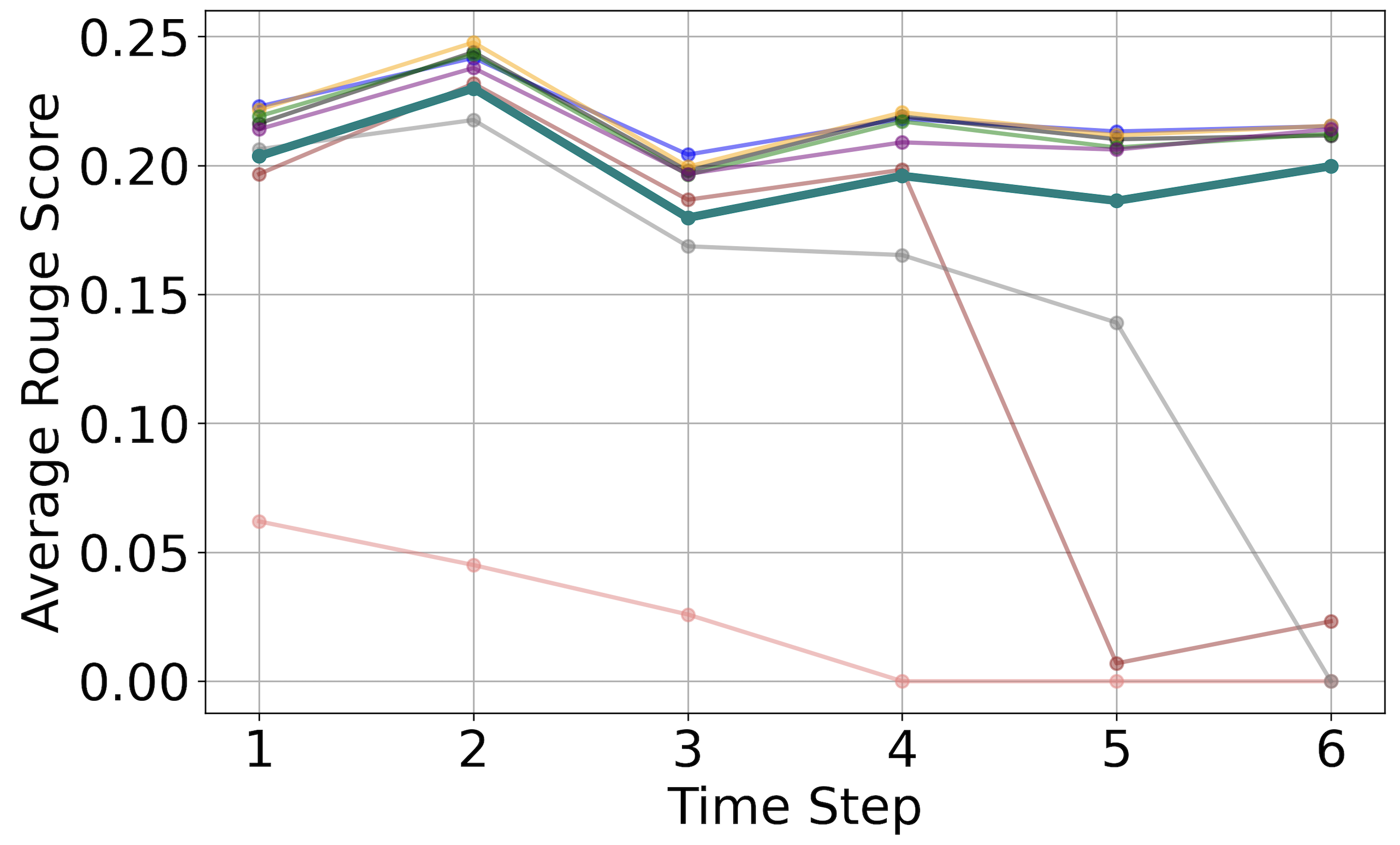}
        \subcaption{Avg Rouge Score on $D_f$}
        \label{fig:main_book_forget_unlearn_mistral}
        \end{subfigure}
        \begin{subfigure}{0.48\textwidth}
        \includegraphics[width=\textwidth]{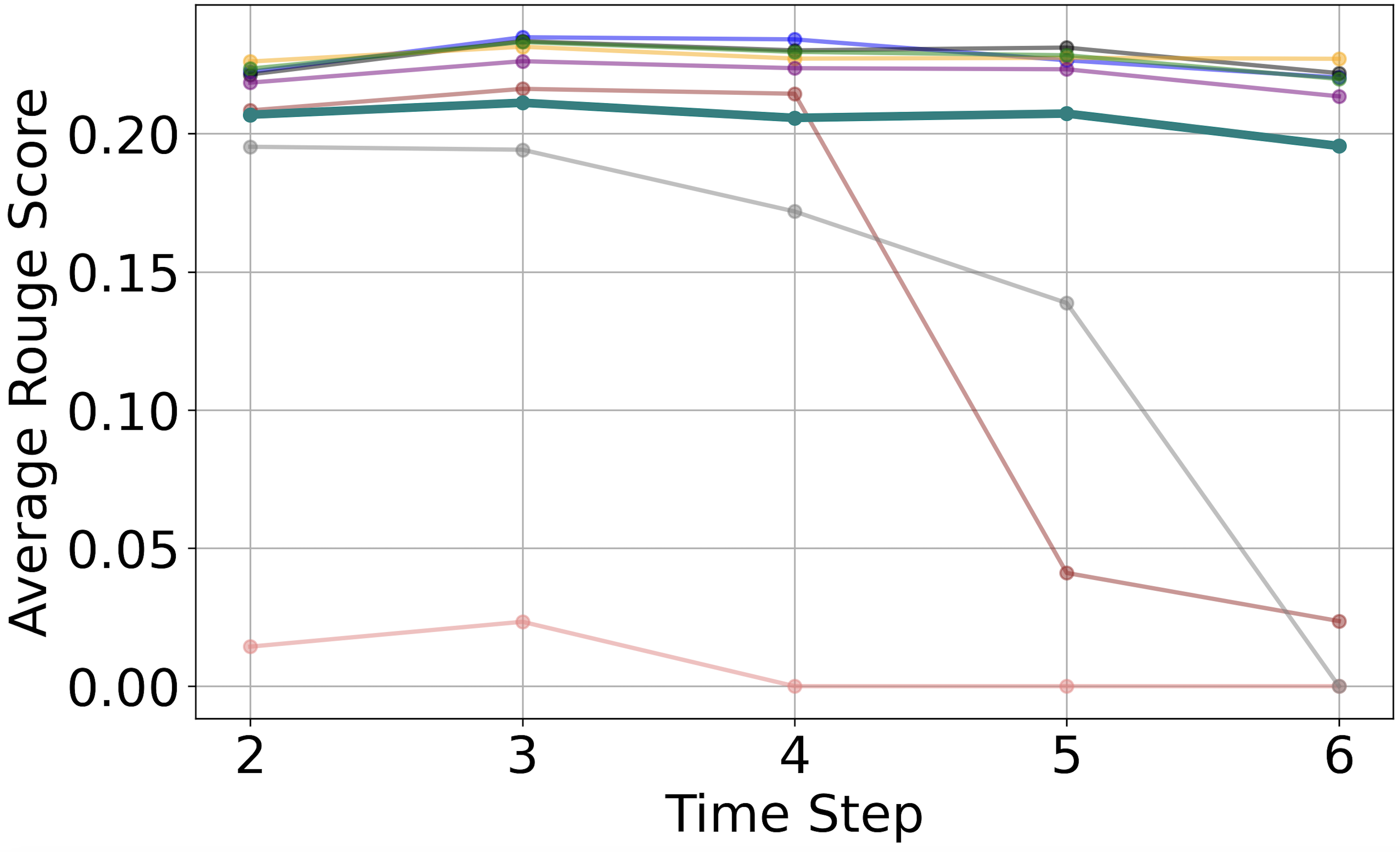}
        \subcaption{Avg Rouge Score on $D_{prev}$}
        \label{fig:main_book_forget_previous_mistral}
        \end{subfigure} 
        
      \begin{subfigure}{0.49\textwidth}
        \includegraphics[width=\textwidth]{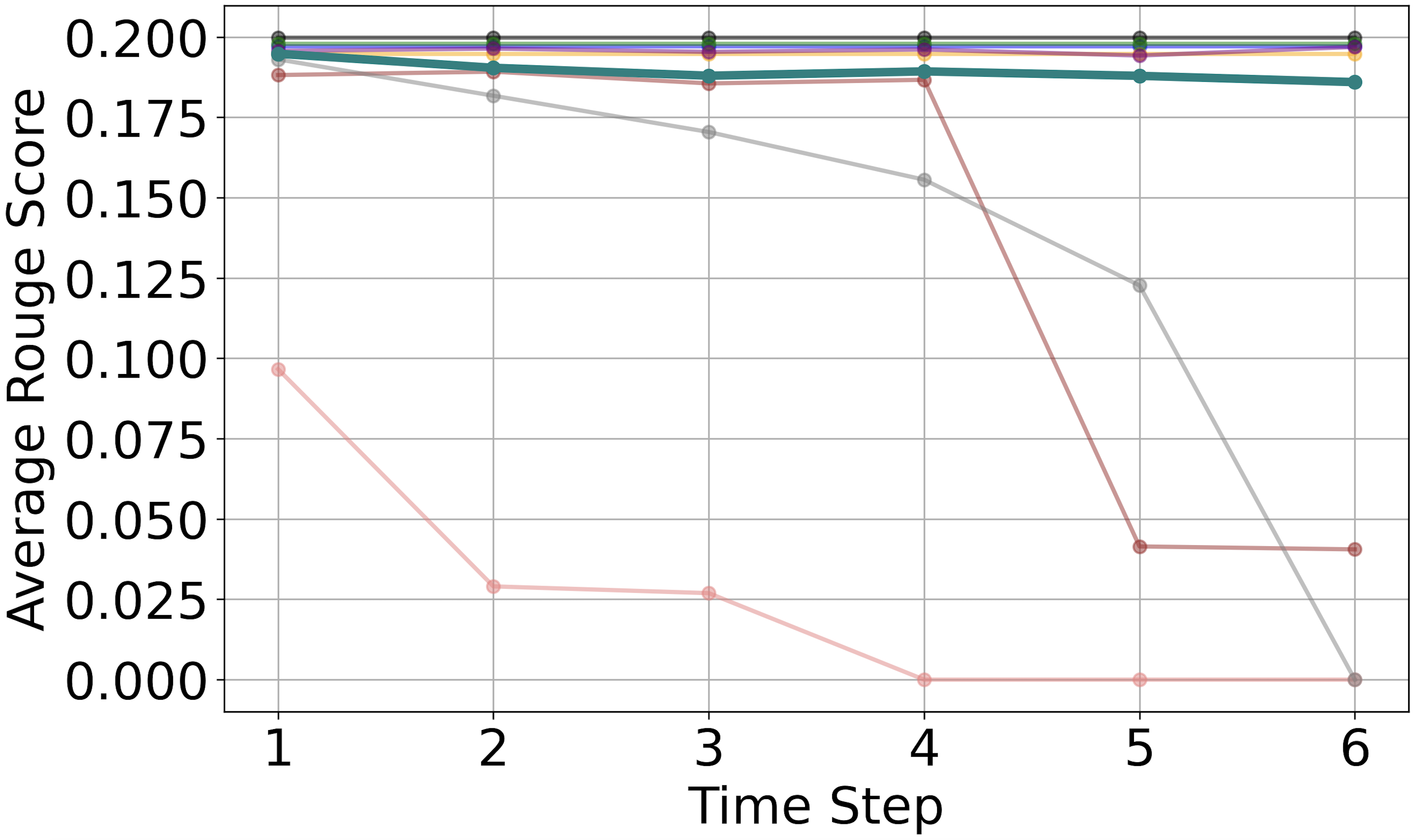}
        \subcaption{Avg Rouge Score on $D_n$}
        \label{fig:main_book_forget_normal_mistral}
        \end{subfigure}
        \begin{subfigure}{0.47\textwidth}
        \includegraphics[width=\textwidth]{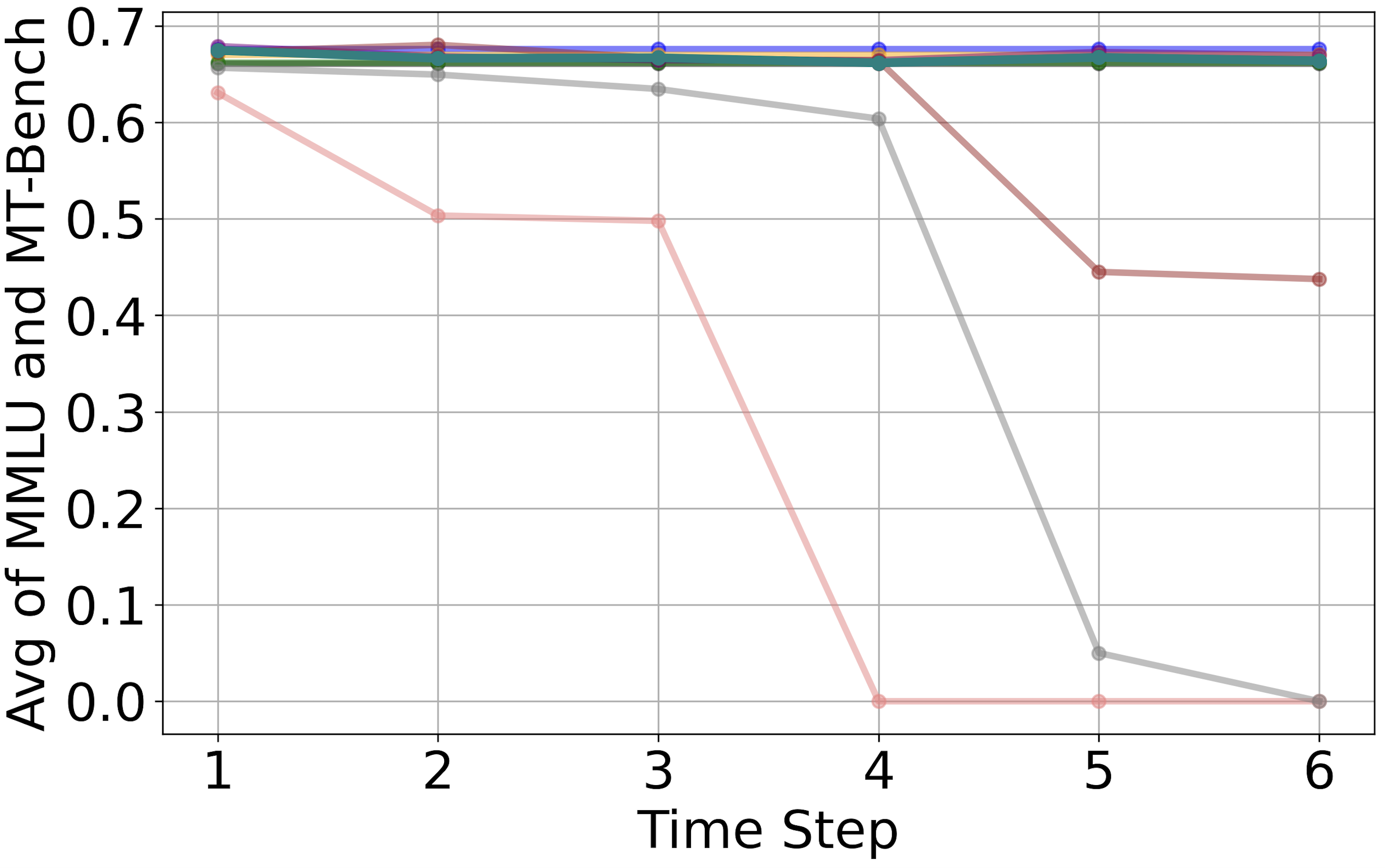}
        \subcaption{Avg MMLU and MT-Bench score}
        \label{fig:main_book_forget_general_mistral}
        \end{subfigure}
\caption{The average of Rouge-1 and Rouge-l score and reasoning abilities for Mistral-7B-Instruct: (a) books to forget $D_f$ ($\downarrow$); (b) previously unlearned books $D_{prev}$ ($\downarrow$); (c) $D_{nor}$ ($\uparrow$). and (d) averaged normalized MMLU and MT-Bench scores ($\uparrow$). The results for TV after time step 8 are omitted due to collapse. Lower Rouge scores for $D_f$ and $D_{prev}$ indicate better unlearning, while higher scores for $D_{nor}$ and benchmarks reflect better performance.}
\label{fig:main_book_forget_all_mistral}
\end{figure*}

\subsection{Non-Targeted Knowledge Retention}
\label{sec:results_knowledge_retention}

This section examines the impact of unlearning on \( D_{nor} \), the books not intended to be unlearned. Ideally, the model should maintain the average Rouge scores compared to the vanilla model.

\paragraph{\method better preserves non-targeted knowledge compared to other unlearning methods.} The results for Llama3.1 and Mistral-7B are notably different. For Mistral-7B, as shown in Figure \ref{fig:main_book_forget_normal_mistral}, the average Rouge scores for GA, Gradient Difference, and TV demonstrate significant loss of retained knowledge at later time steps. While NPO is less effective at mitigating copyright infringement, it consistently retains more knowledge of non-targeted books than \method. In contrast, the results for Llama3.1 fluctuate. As shown in Figure \ref{fig:main_book_forget_normal}, Gradient Difference and \method retain more knowledge of $D_{nrr}$ than the vanilla model. The unexpected re-emergence of these knowledge after unlearning is possibly due to knowledge redistribution during the unlearning process. \cite{yang2024reawakening} also examined this anticipatory recovery behavior where LLMs recover knowledge from the forgetting on documents before encountering them again. Further research is needed to explore this phenomenon. Nevertheless, \method demonstrates stability by preserving knowledge in $D_{nor}$. 

\paragraph{Prompting and MemFree decoding maintain non-targeted knowledge retention.} As shown in Figures \ref{fig:main_book_forget_normal} and \ref{fig:main_book_forget_normal_mistral},
prompting and MemFree consistently retain non-targeted knowledge. Notably, for Llama3.1, both prompting (a) and MemFree retain more knowledge than the vanilla model. In contrast, for Mistral-7B, both prompting methods and MemFree exhibit slightly lower levels of knowledge retention compared to the vanilla model.

\subsection{General-purpose Language Abilities}

\paragraph{GA, Gradient Difference, and TV experienced catastrophic collapse at later time steps.} Figures \ref{fig:main_book_forget_general_llama3.1} and \ref{fig:main_book_forget_general_mistral}, show that GA, Gradient Difference, and TV both have sudden significant degradation of general-purpose abilities at certain time steps. NPO consistently underperforms \method for Llama3.1 outperforms for Mistral-7B. Prompting and MemFree maintain stable general-purpose language abilities, close to the vanilla model. This again demonstrates how existing methods fail to control over weight adjustments during the sequential unlearning process, and how a targeted control becomes crucial. 

\subsection{Copyright Takedown Trade-offs}

We present the overall trade-off of each method in Figures \ref{fig:trade_off_llama3.1} and \ref{fig:trade_off_mistral}. The $x$-axis represents unlearning efficacy score, and the $y$-axis represents general-purpose language ability. 

Specifically, the unlearning efficacy score is defined as:

\begin{equation}
\text{Unlearning Efficacy} = U_f + U_{\text{prev}}
\end{equation}

where 
\begin{equation*}
    U_f = -\frac{1}{2} \left( R_1^f + R_L^f \right)
\end{equation*}
 and \begin{equation*}
 U_{\text{prev}} = -\frac{1}{2} \left( R_1^{\text{prev}} + R_L^{\text{prev}} \right)
 \end{equation*}
 
Here, $R_1^f$ and $R_L^f$ represent the Rouge-1 and Rouge-L scores for $D_f$, and $R_1^{\text{prev}}$ and $R_L^{\text{prev}}$ represent the Rouge-1 and Rouge-L scores for $D_{prev}$.

The general-purpose language ability is defined as:
\begin{equation}
\text{General-purpose Language Ability} = \frac{1}{2} \left( M + \frac{B}{10} \right)
\end{equation}

where $M$ and $B$ represent the MMLU and MT-Bench scores, respectively.

An ideal copyright takedown method would aim for the top-right corner. Comparing to existing baseline methods across two different models, \textbf{\method achieves a better trade-off between unlearning efficacy and general-purpose language abilities retention.} 

\begin{figure*}
\centering
         \begin{subfigure}[t]{\textwidth}
            \centering
            \includegraphics[width=1\textwidth]{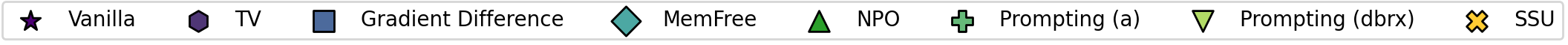}
          \end{subfigure}
        \begin{subfigure}{0.44\textwidth}
        \includegraphics[width=\textwidth]{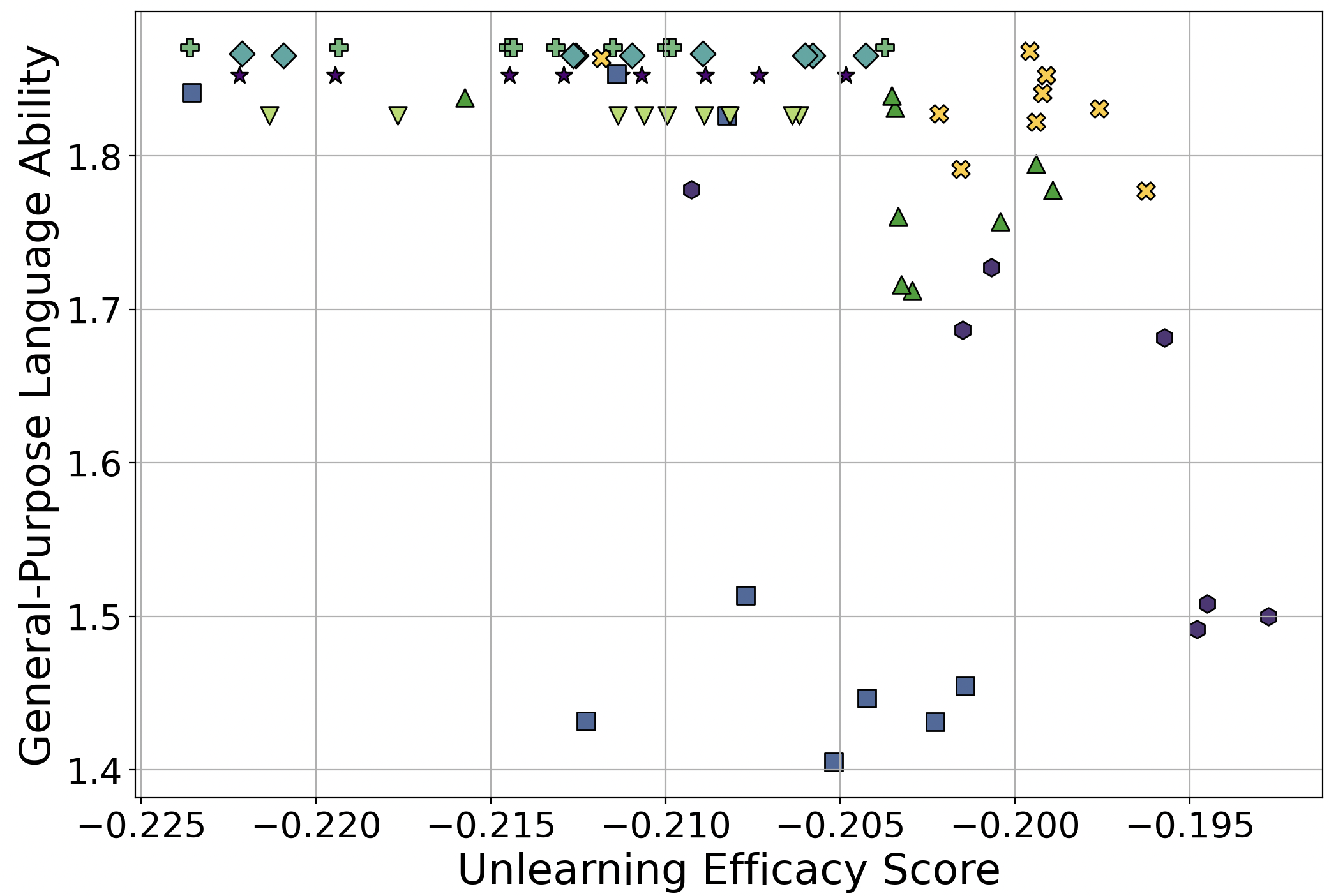}
        \subcaption{Llama3.1-8B-Instruct Trade-off}
        \label{fig:trade_off_llama3.1}
        \end{subfigure}
        \begin{subfigure}{0.45\textwidth}
        \includegraphics[width=\textwidth]{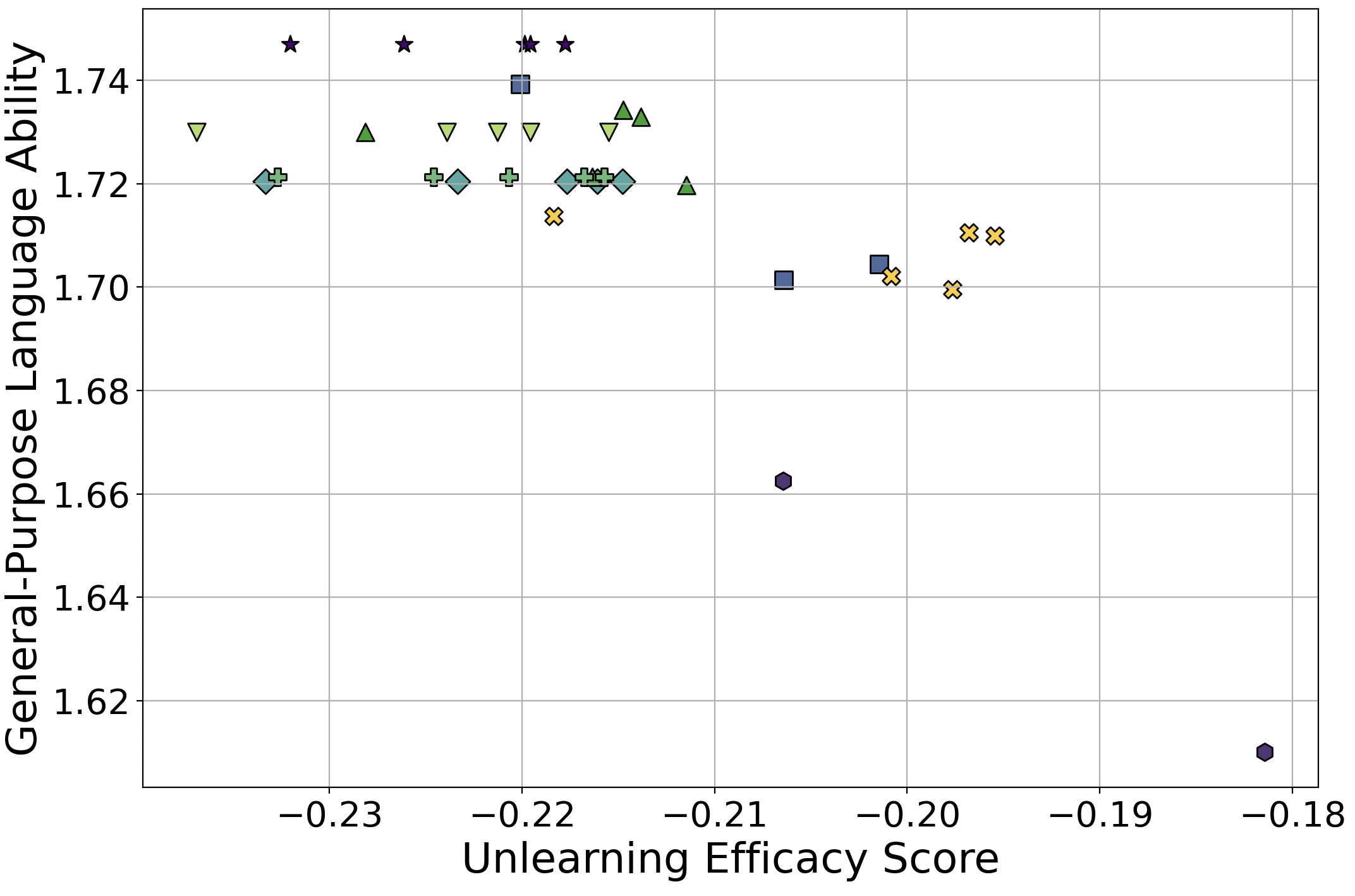}
        \subcaption{Mistral-7B-Instruct Trade-off}
        \label{fig:trade_off_mistral}
        \end{subfigure} 
\caption{Trade-off between general-purpose language abilities and unlearning efficacy for Llama3.1 and Mistral-7B, including all methods, except TV beyond time step 9 (Llama3.1) and time step 3 (Mistral-7B), and Gradient Difference beyond time step 4 (Mistral-7B), as they all collapsed during these time steps. General-purpose abilities are represented by the average of MMLU and MT-Bench scores, normalized. Unlearning efficacy is measured as the average of Rouge-1 and Rouge-L scores on $D_f$ and $D_{prev}$, where lower Rouge scores indicate better unlearning performance; thus, values were negated for clarity. The ideal performance is positioned in the top-right corner. The plots capture the performance of all methods at every time step greater than 1.}
\label{fig:trade_off_appendix}
\end{figure*}

\begin{figure*}[th]
\centering
         \begin{subfigure}[b]{\textwidth}
            \centering
            \includegraphics[width=0.45\textwidth]{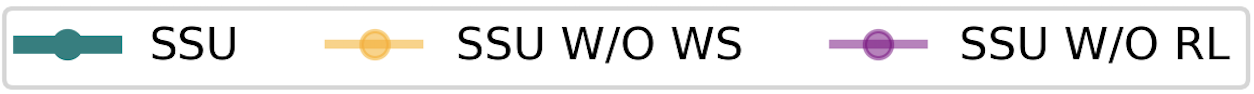}
          \end{subfigure}
        \begin{subfigure}{0.24\textwidth}
        \includegraphics[width=\textwidth]{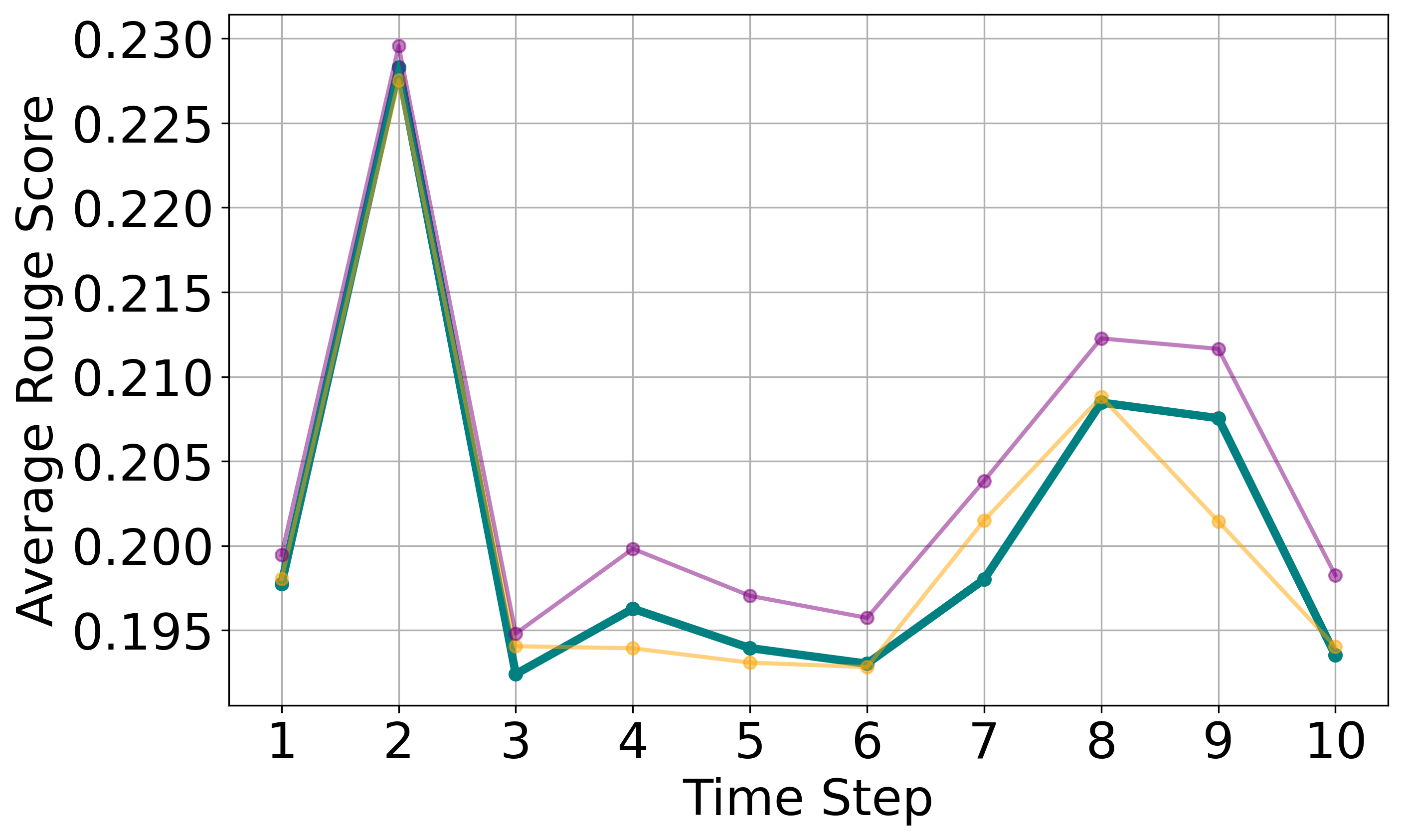}
        \subcaption{Avg Rouge on $D_f$}
        \label{fig:ablation_book_forget_llama3.1}
        \end{subfigure}
        \begin{subfigure}{0.25\textwidth}
        \includegraphics[width=\textwidth]{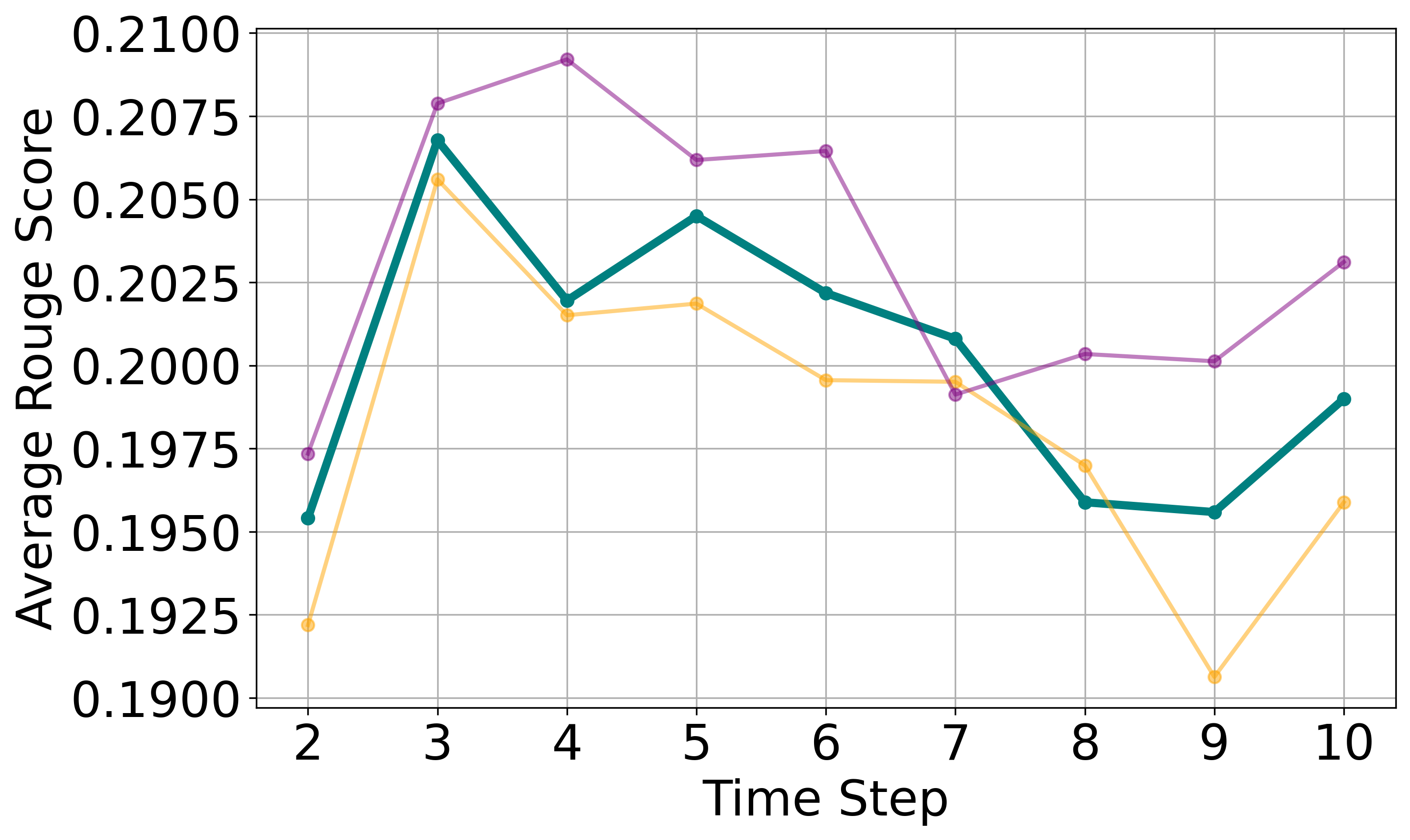}
        \subcaption{Avg Rouge on $D_{prev}$}
        \label{fig:ablation_book_prev_llama3.1}
        \end{subfigure}
        \begin{subfigure}{0.24\textwidth}
        \includegraphics[width=\textwidth]{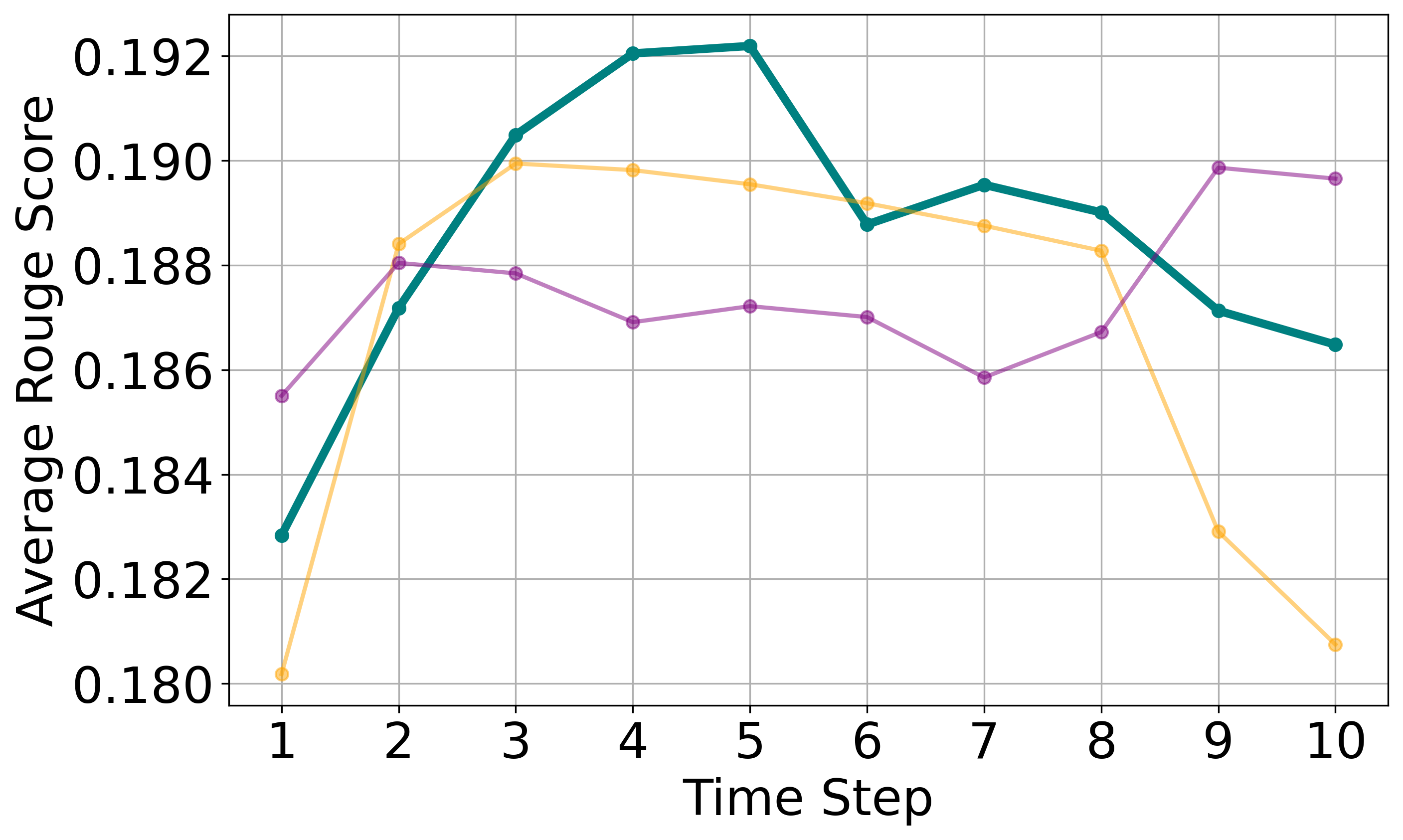}
        \subcaption{Avg Rouge on $D_n$}
        \label{fig:ablation_book_norm_llama3.1)}
        \end{subfigure}
        \begin{subfigure}{0.24\textwidth}
        \includegraphics[width=\textwidth]{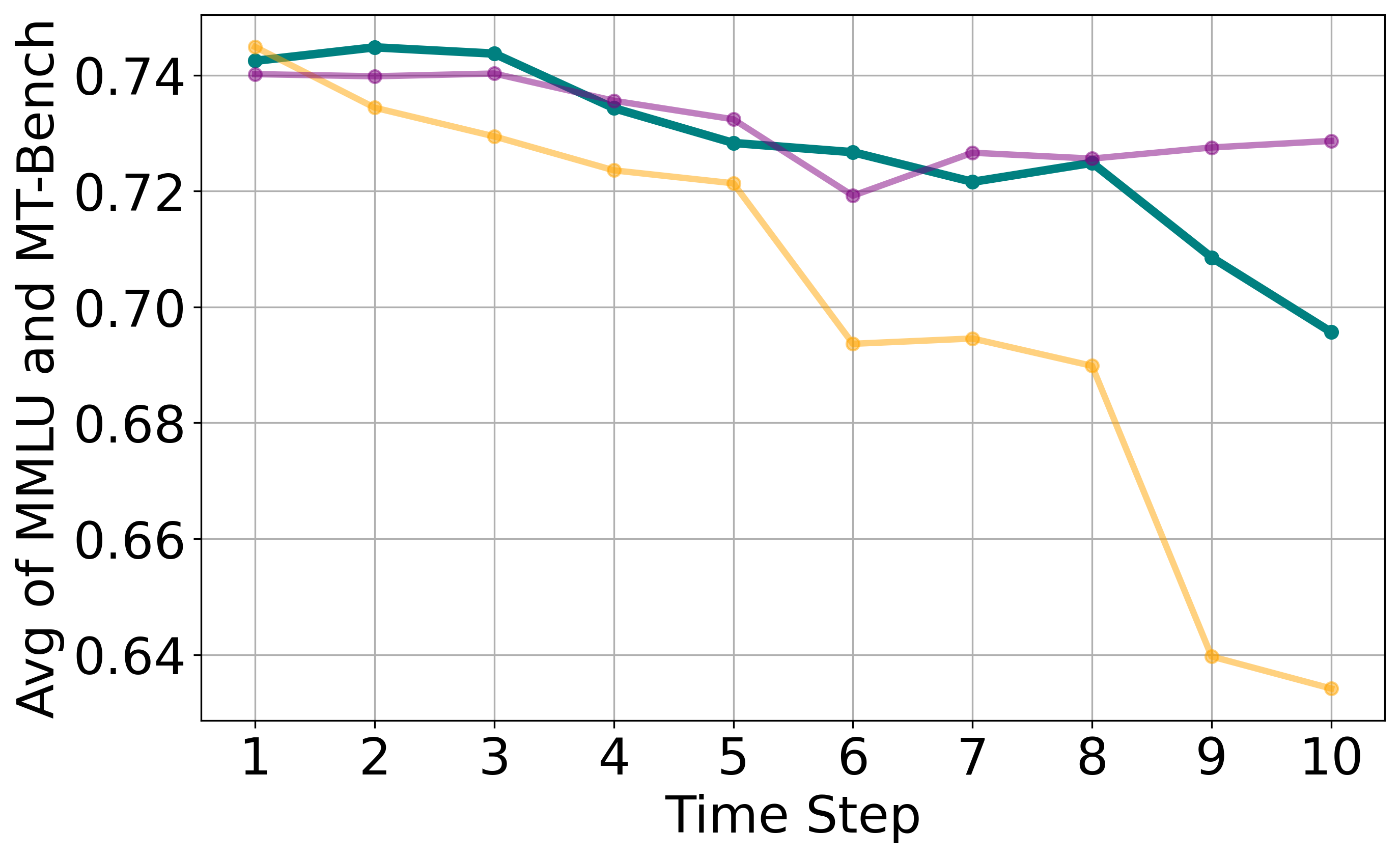}
        \subcaption{General Performance}
        \label{fig:ablation_general_llama3.1)}
        \end{subfigure}
\caption{Ablation study of \method for Llama3.1-8B-Instruct. The orange line represents unlearning without the weight saliency map, while the purple line shows the effect of removing the random labeling loss.}
\label{fig:ablation_llama}
\end{figure*}

\begin{figure*}[btp]
\centering
         \begin{subfigure}[b]{\textwidth}
            \centering
            \includegraphics[width=0.45\textwidth]{Figure/ablation/legend_ablation_plog.png}
          \end{subfigure}
        \begin{subfigure}{0.24\textwidth}
        \includegraphics[width=\textwidth]{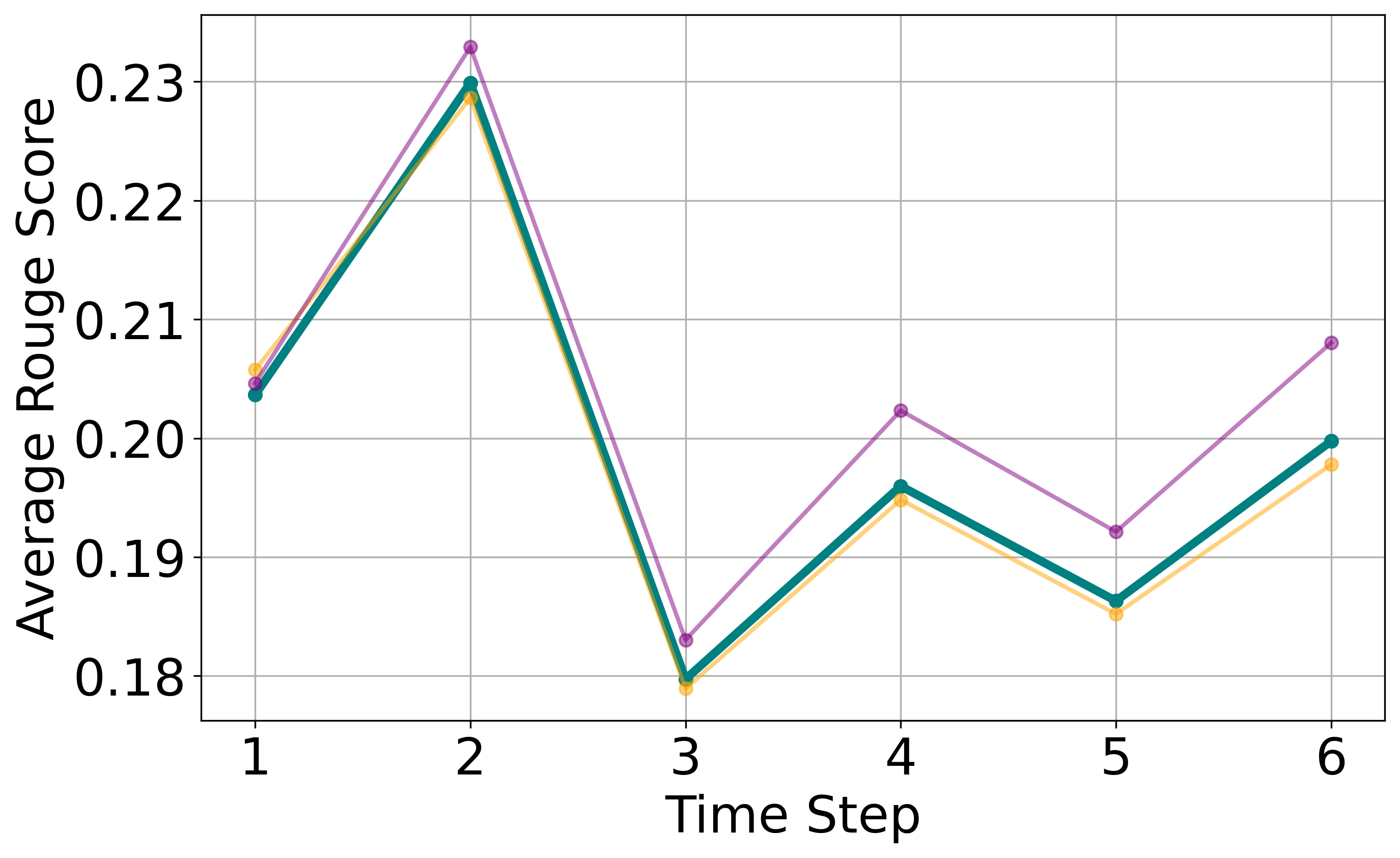}
        \subcaption{Avg Rouge on $D_f$}
        \label{fig:ablation_book_forget_mistral}
        \end{subfigure}
        \begin{subfigure}{0.24\textwidth}
        \includegraphics[width=\textwidth]{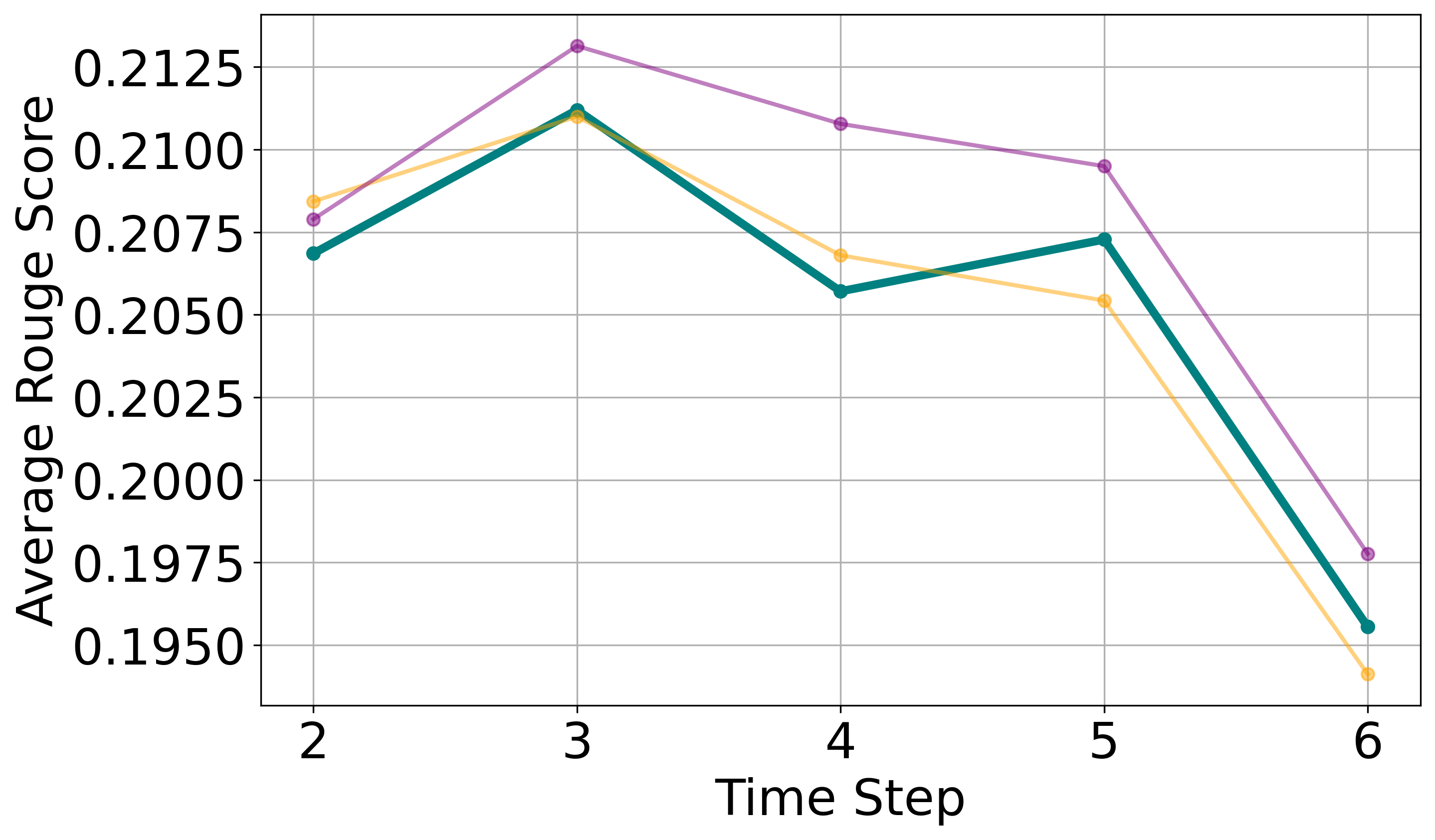}
        \subcaption{Avg Rouge on $D_{prev}$}
        \label{fig:ablation_book_prev_mistral}
        \end{subfigure}
        \begin{subfigure}{0.24\textwidth}
        \includegraphics[width=\textwidth]{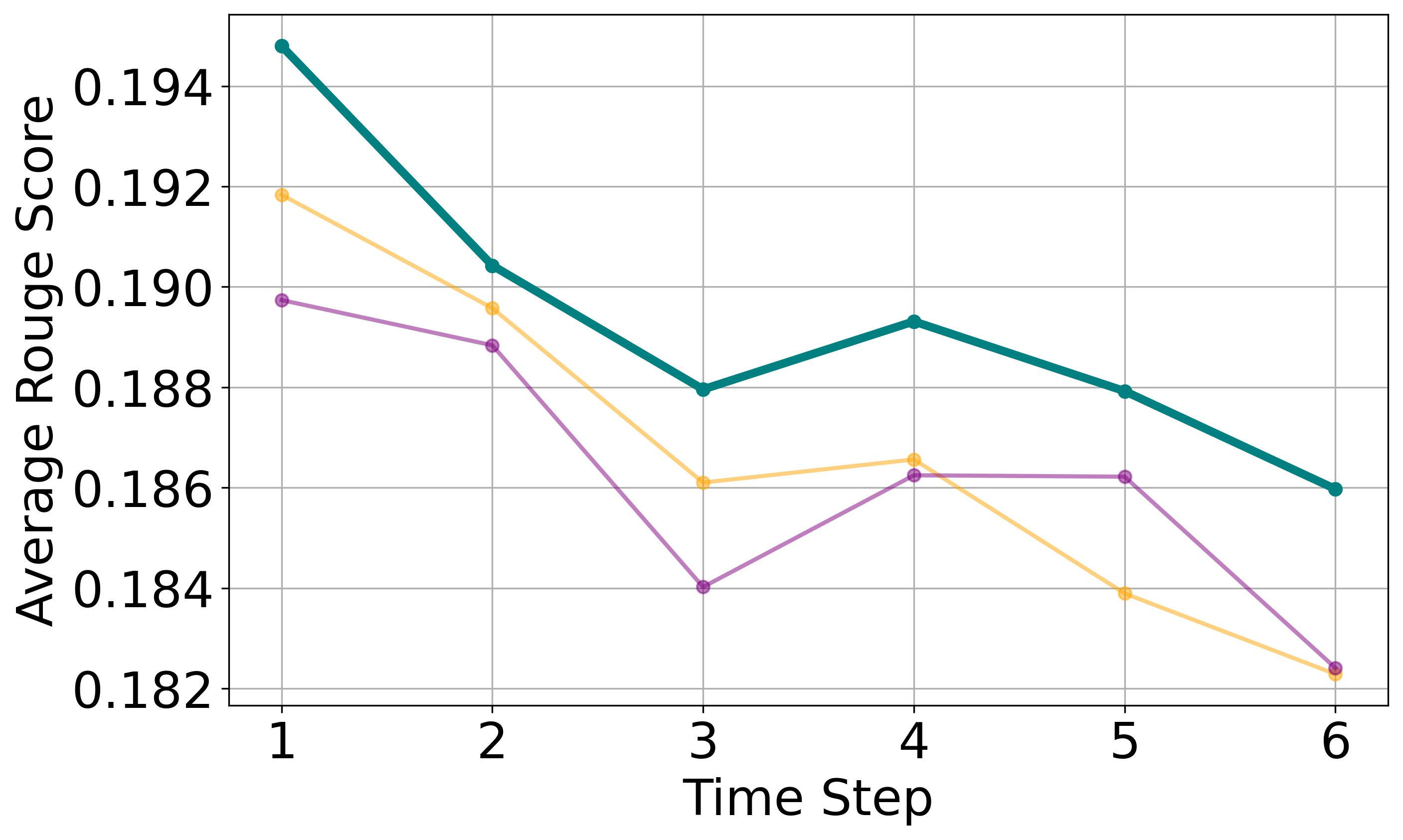}
        \subcaption{Avg Rouge on $D_n$}
        \label{fig:ablation_book_norm_mistral)}
        \end{subfigure}
        \begin{subfigure}{0.24\textwidth}
        \includegraphics[width=\textwidth]{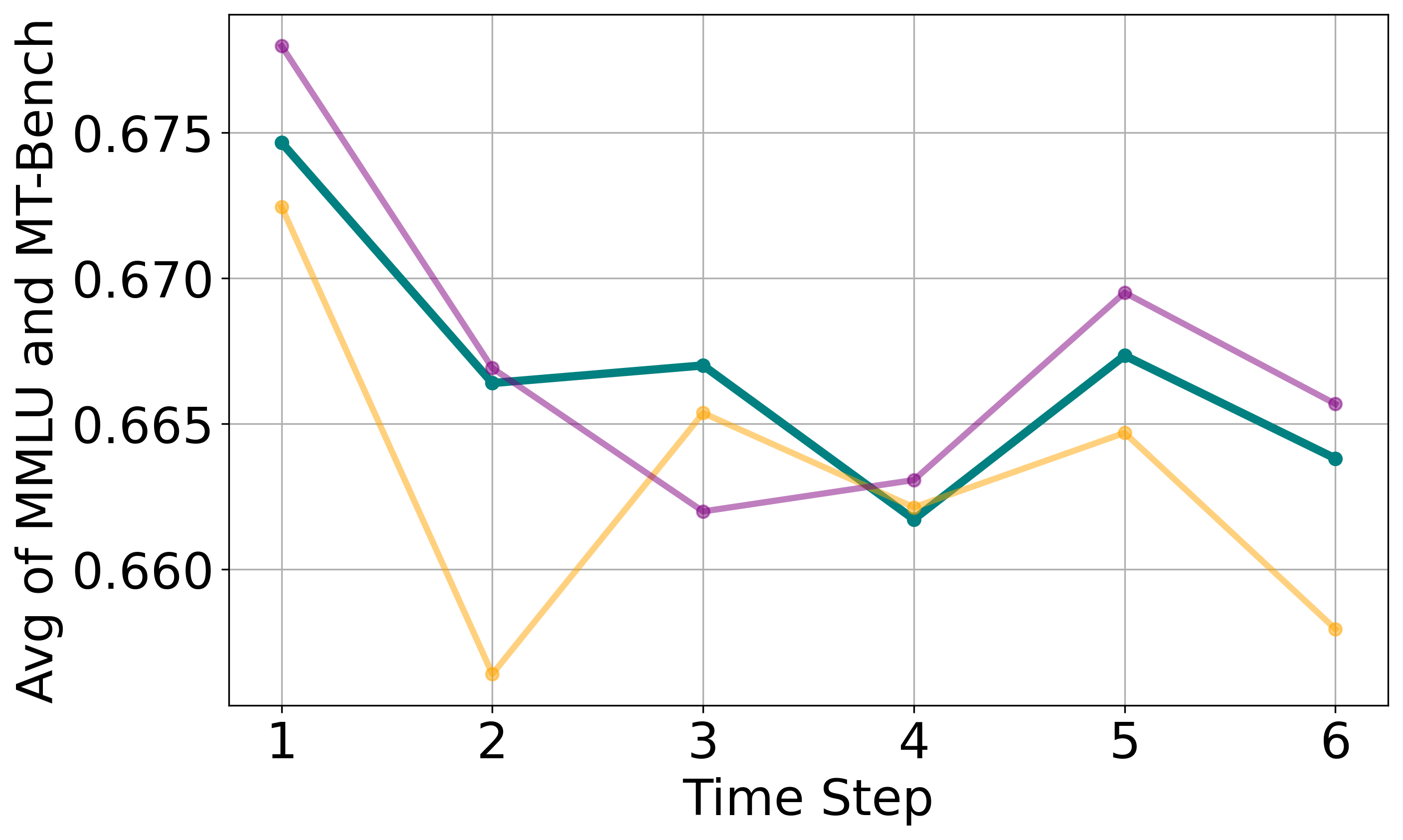}
        \subcaption{General Performance}
        \label{fig:ablation_general_mistral)}
        \end{subfigure}
\caption{Ablation study of \method for Mistral-7B-Instruct-v0.3. The orange line represents unlearning without the weight saliency map, while the purple line shows the effect of removing the random labeling loss.}
\label{fig:ablation_mistral}
\end{figure*}

\section{Ablation Study} 
\vspace{-0.1in}
We analyze the impact of different components of \method, including weight saliency maps and random labeling loss, on the sequential unlearning process. Figure \ref{fig:ablation_llama} presents results for Llama3.1-8B-Instruct, and Figure \ref{fig:ablation_mistral} shows ablation results for Mistral-7B-Instruct-v0.3.

\subsection{Impact of Weight Saliency}
As shown in Figures  \ref{fig:ablation_book_forget_llama3.1} and \ref{fig:ablation_book_prev_llama3.1}, the performance of \method without weight saliency leads to lower average Rouge scores on both $D_f$ and $D_{nor}$. Additionally, Figure \ref{fig:ablation_general_llama3.1)} shows a sharper decline in benchmark performance with each time step, indicating that without weight saliency, the risk of catastrophic collapse increases as general-purpose language abilities deteriorate. \textbf{By updating only specific parts of the model's weights, weight saliency helps preserve the general-purpose language abilities.}

\subsection{Impact of Random Labeling Loss}
As seen in Figures \ref{fig:ablation_book_forget_llama3.1} and \ref{fig:ablation_book_prev_llama3.1}, \method without random labeling loss results in a higher average Rouge scores and slightly improved general-purpose language abilities. This suggests that \textbf{random labeling loss enhances the model’s ability to unlearn $D_f$ consistently across all time steps.}

It is also noteworthy that, as shown in Figures \ref{fig:ablation_book_norm_llama3.1)} and \ref{fig:ablation_book_norm_mistral)}, \method without weight saliency and \method without random loss lead to lower average Rouge scores for $D_{nor}$ in both models. This observation highlights the need for further research on the impact of unlearning algorithms on non-targeted knowledge retention, particularly in preserving content not intended for unlearning.

\section{Numeric Experiment Results}
In this section, we present our experimental results numerically. Tables \ref{tab:main_results_time_step_1} to \ref{tab:main_results_time_step_10} display the unlearning results of Llama3.1 across all ten time steps. Tables \ref{tab:main_results_mistral_time_step_1} to \ref{tab:main_results_mistral_time_step_6} show the results for Mistral-7B unlearning across all six time steps. In sections \ref{section:appendix-collapse-llama3.1} and \ref{section:appendix-collapse-mistral7B} we illustrate how GA, TV, and Gradient Difference encounter catastrophic collapse. 

\subsection{Catastrophic Collapse of Llama3.1}
\label{section:appendix-collapse-llama3.1}

\textbf{GA, TV, and Gradient Difference experience varying levels of model collapse during the sequential unlearning process.} 
As shown in table \ref{tab:main_results_time_step_1}, GA begins with an MMLU of 0.5821. By time step 5, GA's MMLU has dropped to 0.3102 \ref{tab:main_results_time_step_5}, demonstrating a rapid degradation in general reasoning. Similarly, TV starts with an MMLU of 0.6621 at time step 1 and undergoes a steep decline in reasoning ability, dropping to 0.4887 by time step 5, and reaching an MMLU of 0 at time step 10 (as shown in tables \ref{tab:main_results_time_step_5} and \ref{tab:main_results_time_step_10}). Additionally, Gradient Difference also faces catastrophic collapse at time step 5. Specifically, its MT-Bench score falls from 8.034 at time step 4 (table \ref{tab:main_results_time_step_4}) to 4.9438 at time step 5, and further declines to 4.48 at time step 10, though its MMLU score remains stable. Lastly, NPO and \method exhibit a gradual decline in general-purpose language abilities, but \method consistently outperforms NPO across all the time steps. 

\subsection{Catastrophic Collapse of Mistral-7B}
\label{section:appendix-collapse-mistral7B}
\textbf{GA and TV experience rapid model collapse, while Gradient Difference still suffers loss of conversational ability.} 
As shown in table \ref{tab:main_results_mistral_time_step_1}, GA collapses at the first time step. TV initially declines gradually but undergoes a sudden reasoning degradation at time step 5 (table \ref{tab:main_results_mistral_time_step_4}), where both MMLU and MT-Bench scores drop to 0. Gradient Difference experiences a sharp decrease in MT-Bench score from 7.2375 at time step 4 to 2.8375 at time step 5, eventually dropping to 2.6815 at time step 6. Similar to the Llama3.1 case, Gradient Difference maintains stable performance on MMLU. Lastly, NPO maintains competitive general-purpose abilities scores compared to \method, Figures \ref{fig:main_book_forget_all_mistral} and \ref{fig:trade_off_appendix} show that \method demonstrates superior unlearning efficacy and achieves a better trade-off. 

\begin{table*}[h]
\centering
\resizebox{1\textwidth}{!}{
\begin{tabular}{l||ll||ll||ll||ll}
\hline
\multicolumn{1}{c||}{\multirow{2}{*}{}} & 
\multicolumn{2}{c||}{\textbf{$\mathbf{D_f}$}} & 
\multicolumn{2}{c||}{\textbf{$\mathbf{D_{prev}}$}} & 
\multicolumn{2}{c||}{\textbf{$\mathbf{D_{nor}}$}} &
\multicolumn{2}{c}{\textbf{Benchmark}} \\ \cline{2-9}

\multicolumn{1}{c||}{} &
\begin{tabular}[c]{@{}l@{}}Rouge-1 \\ \end{tabular} &
\begin{tabular}[c]{@{}l@{}}Rouge-L\\ \end{tabular} &
\begin{tabular}[c]{@{}l@{}}Rouge-1\\ \end{tabular} &
\begin{tabular}[c]{@{}l@{}}Rouge-L \\  \end{tabular} &
\begin{tabular}[c]{@{}l@{}}Rouge-1 \\ \end{tabular} &
\begin{tabular}[c]{@{}l@{}}Rouge-L \\  \end{tabular} &
\begin{tabular}[c]{@{}l@{}}MMLU \\ \end{tabular} &
\begin{tabular}[c]{@{}l@{}}MT-Bench \\ \end{tabular} 
\\ \hline
Vanilla &  0.2724 & 0.1530 & 0 & 0 & 0.2349 & 0.1380 & 0.6618 & 8.1808 \\
Prompting (a) & 0.2707 & 0.1541 & 0 & 0 & 0.2376 & 0.1364 & 0.6635 & 8.3344 \\
Prompting (dbrx) & 0.2730 & 0.1535 & 0 & 0 & 0.2333 & 0.1364 & 0.6611 & 7.9563 \\
MemFree Decode & 0.2711 & 0.1524 & 0 & 0 & 0.2392 & 0.1407 & 0.6618 & 8.2453 \\
GA & 0.2504 & 0.1430 & 0 & 0 & 0.2282 & 0.1354 & 0.5821 & 8.1719 \\
NPO & 0.2655 & 0.1487 & 0 & 0 & 0.2380 & 0.1411 & 0.6600 & 8.1938 \\
Gradiet Difference & 0.2619 & 0.1496 & 0 & 0 & 0.2433 & 0.1241 & 0.6544 & 8.0031 \\
TV & 0.2463 & 0.1433 & 0 & 0 & 0.2228 & 0.1331 & 0.6621 & 8.2038 \\
SSU & 0.2523 & 0.1432 &0 & 0 & 0.2299 & 0.1357 & 0.6625 & 8.2250 \\
\hline
\end{tabular}}
\caption{Overall results of Llama3.1 at time step 1, compared with several baselines for $D_f$, $D_{prev}$, and $D_{nor}$. Benchmark performance includes MMLU and MT-Bench scores.}
\vspace{-0.15in}
\label{tab:main_results_time_step_1}
\end{table*}

\begin{table*}[h]
\centering
\resizebox{1\textwidth}{!}{
\begin{tabular}{l||ll||ll||ll||ll}
\hline
\multicolumn{1}{c||}{\multirow{2}{*}{}} & 
\multicolumn{2}{c||}{\textbf{$\mathbf{D_f}$}} & 
\multicolumn{2}{c||}{\textbf{$\mathbf{D_{prev}}$}} & 
\multicolumn{2}{c||}{\textbf{$\mathbf{D_{nor}}$}} &
\multicolumn{2}{c}{\textbf{Benchmark}} \\ \cline{2-9}

\multicolumn{1}{c||}{} &
\begin{tabular}[c]{@{}l@{}}Rouge-1 \\ \end{tabular} &
\begin{tabular}[c]{@{}l@{}}Rouge-L\\ \end{tabular} &
\begin{tabular}[c]{@{}l@{}}Rouge-1\\ \end{tabular} &
\begin{tabular}[c]{@{}l@{}}Rouge-L \\  \end{tabular} &
\begin{tabular}[c]{@{}l@{}}Rouge-1 \\ \end{tabular} &
\begin{tabular}[c]{@{}l@{}}Rouge-L \\  \end{tabular} &
\begin{tabular}[c]{@{}l@{}}MMLU \\ \end{tabular} &
\begin{tabular}[c]{@{}l@{}}MT-Bench \\ \end{tabular} 
\\ \hline
Vanilla &  0.3027 & 0.1793 & 0.2586 & 0.1478 & 0.2349 & 0.1380 & 0.6618 & 8.1808 \\
Prompting (a) & 0.3036 & 0.1814 & 0.2605 & 0.1489 & 0.2376 & 0.1364 & 0.6635 & 8.3344 \\
Prompting (dbrx) & 0.2971 & 0.1759 & 0.2638 & 0.1485 & 0.2333 & 0.1364 & 0.6611 & 7.9563 \\
MemFree Decode & 0.2995 & 0.1781 & 0.2604 & 0.1501 & 0.2392 & 0.1407 & 0.6618 & 8.2453 \\
GA & 0.2755 & 0.1644 & 0.2489 & 0.1409 & 0.2287 & 0.1375 & 0.5157 & 8.1719 \\
NPO & 0.2910 & 0.1725 & 0.2556 & 0.1439 & 0.2368 & 0.1395 & 0.6512 & 8.1063 \\
Gradient Difference & 0.2996 & 0.1806 & 0.2605 & 0.1494 & 0.2401 & 0.1405 & 0.6544 & 8.0627 \\
TV & 0.2774 & 0.1674 & 0.2492 & 0.1433 & 0.2425 & 0.1431 & 0.5845 & 8.0808 \\
SSU & 0.2863 & 0.1702 & 0.2494 & 0.1414 & 0.2354 & 0.1390 & 0.6519 & 8.3769 \\
\hline
\end{tabular}}
\caption{Overall results of Llama3.1 at time step 2, compared with several baselines for $D_f$, $D_{prev}$, and $D_{nor}$. Benchmark performance includes MMLU and MT-Bench scores.}
\vspace{-0.15in}
\label{tab:main_results_time_step_2}
\end{table*}

\begin{table*}[h]
\centering
\resizebox{1\textwidth}{!}{
\begin{tabular}{l||ll||ll||ll||ll}
\hline
\multicolumn{1}{c||}{\multirow{2}{*}{}} & 
\multicolumn{2}{c||}{\textbf{$\mathbf{D_f}$}} & 
\multicolumn{2}{c||}{\textbf{$\mathbf{D_{prev}}$}} & 
\multicolumn{2}{c||}{\textbf{$\mathbf{D_{nor}}$}} &
\multicolumn{2}{c}{\textbf{Benchmark}} \\ \cline{2-9}

\multicolumn{1}{c||}{} &
\begin{tabular}[c]{@{}l@{}}Rouge-1 \\ \end{tabular} &
\begin{tabular}[c]{@{}l@{}}Rouge-L\\ \end{tabular} &
\begin{tabular}[c]{@{}l@{}}Rouge-1\\ \end{tabular} &
\begin{tabular}[c]{@{}l@{}}Rouge-L \\  \end{tabular} &
\begin{tabular}[c]{@{}l@{}}Rouge-1 \\ \end{tabular} &
\begin{tabular}[c]{@{}l@{}}Rouge-L \\  \end{tabular} &
\begin{tabular}[c]{@{}l@{}}MMLU \\ \end{tabular} &
\begin{tabular}[c]{@{}l@{}}MT-Bench \\ \end{tabular} 
\\ \hline
Vanilla & 0.2544 & 0.1534 & 0.2756 & 0.1617 & 0.2349 & 0.1380 & 0.6618 & 8.1808 \\
Prompting (a) & 0.2568 & 0.1556 & 0.2812 & 0.1638 & 0.2376 & 0.1364 & 0.6635 & 8.3344 \\
Prompting (dbrx) & 0.2464 & 0.1496 & 0.2778 & 0.1618 & 0.2333 & 0.1364 & 0.6611 & 7.9563 \\
MemFree Decode & 0.2456 & 0.1508 & 0.2773 & 0.1621 & 0.2392 & 0.1407 & 0.6618 & 8.2453 \\
GA & 0.2459 & 0.1483 & 0.2569 & 0.1527 & 0.2294 & 0.1353 & 0.4940 & 7.8031 \\
NPO & 0.2466 & 0.1492 & 0.2653 & 0.1527 & 0.2375 & 0.1401 & 0.6481 & 8.0547 \\
Gradient Difference & 0.2493 & 0.1523 & 0.2737 & 0.1577 & 0.2409 & 0.1417 & 0.6544 & 7.9727 \\
TV & 0.2439 & 0.1473 & 0.2587 & 0.1527 & 0.2414 & 0.1430 & 0.5316 & 8.1163 \\
SSU & 0.2391 & 0.1458 & 0.2627 & 0.1509 & 0.2398 & 0.1413 & 0.6481 & 8.3938 \\
\hline
\end{tabular}}
\caption{Overall results of Llama3.1 at time step 3, compared with several baselines for $D_f$, $D_{prev}$, and $D_{nor}$. Benchmark performance includes MMLU and MT-Bench scores.}
\vspace{-0.15in}
\label{tab:main_results_time_step_3}
\end{table*}

\begin{table*}[h]
\centering
\resizebox{1\textwidth}{!}{
\begin{tabular}{l||ll||ll||ll||ll}
\hline
\multicolumn{1}{c||}{\multirow{2}{*}{}} & 
\multicolumn{2}{c||}{\textbf{$\mathbf{D_f}$}} & 
\multicolumn{2}{c||}{\textbf{$\mathbf{D_{prev}}$}} & 
\multicolumn{2}{c||}{\textbf{$\mathbf{D_{nor}}$}} &
\multicolumn{2}{c}{\textbf{Benchmark}} \\ \cline{2-9}

\multicolumn{1}{c||}{} &
\begin{tabular}[c]{@{}l@{}}Rouge-1 \\ \end{tabular} &
\begin{tabular}[c]{@{}l@{}}Rouge-L\\ \end{tabular} &
\begin{tabular}[c]{@{}l@{}}Rouge-1\\ \end{tabular} &
\begin{tabular}[c]{@{}l@{}}Rouge-L \\  \end{tabular} &
\begin{tabular}[c]{@{}l@{}}Rouge-1 \\ \end{tabular} &
\begin{tabular}[c]{@{}l@{}}Rouge-L \\  \end{tabular} &
\begin{tabular}[c]{@{}l@{}}MMLU \\ \end{tabular} &
\begin{tabular}[c]{@{}l@{}}MT-Bench \\ \end{tabular} 
\\ \hline
Vanilla & 0.2589 & 0.1530 & 0.2721 & 0.1588 & 0.2349 & 0.1380 & 0.6618 & 8.1808 \\
Prompting (a) & 0.2572 & 0.1522 & 0.2756 & 0.1609 & 0.2376 & 0.1364 & 0.6635 & 8.3344 \\
Prompting (dbrx) & 0.2597 & 0.1531 & 0.2713 & 0.1583 & 0.2333 & 0.1364 & 0.6611 & 7.9563 \\
MemFree Decode & 0.2591 & 0.1520 & 0.2722 & 0.1605 & 0.2383 & 0.1408 & 0.6617 & 8.2453 \\
GA & 0.2599 & 0.1466 & 0.2533 & 0.1501 & 0.2295 & 0.1355 & 0.4853 & 7.9813 \\
NPO & 0.2480 & 0.1478 & 0.2641 & 0.1542 & 0.2364 & 0.1407 & 0.6537 & 8.0821 \\
Gradient Difference & 0.2606 & 0.1504 & 0.2748 & 0.1598 & 0.2464 & 0.1457 & 0.6579 & 8.0344 \\
TV & 0.2518 & 0.1500 & 0.2564 & 0.1519 & 0.2342 & 0.1396 & 0.4982 & 8.1456 \\
SSU & 0.2489 & 0.1436 & 0.2522 & 0.1516 & 0.2417 & 0.1424 & 0.6432 & 8.2547 \\
\hline
\end{tabular}}
\caption{Overall results of Llama3.1 at time step 4, compared with several baselines for $D_f$, $D_{prev}$, and $D_{nor}$. Benchmark performance includes MMLU and MT-Bench scores.}
\vspace{-0.15in}
\label{tab:main_results_time_step_4}
\end{table*}

\begin{table*}[h]
\centering
\resizebox{1\textwidth}{!}{
\begin{tabular}{l||ll||ll||ll||ll}
\hline
\multicolumn{1}{c||}{\multirow{2}{*}{}} & 
\multicolumn{2}{c||}{\textbf{$\mathbf{D_f}$}} & 
\multicolumn{2}{c||}{\textbf{$\mathbf{D_{prev}}$}} & 
\multicolumn{2}{c||}{\textbf{$\mathbf{D_{nor}}$}} &
\multicolumn{2}{c}{\textbf{Benchmark}} \\ \cline{2-9}

\multicolumn{1}{c||}{} &
\begin{tabular}[c]{@{}l@{}}Rouge-1 \\ \end{tabular} &
\begin{tabular}[c]{@{}l@{}}Rouge-L\\ \end{tabular} &
\begin{tabular}[c]{@{}l@{}}Rouge-1\\ \end{tabular} &
\begin{tabular}[c]{@{}l@{}}Rouge-L \\  \end{tabular} &
\begin{tabular}[c]{@{}l@{}}Rouge-1 \\ \end{tabular} &
\begin{tabular}[c]{@{}l@{}}Rouge-L \\  \end{tabular} &
\begin{tabular}[c]{@{}l@{}}MMLU \\ \end{tabular} &
\begin{tabular}[c]{@{}l@{}}MT-Bench \\ \end{tabular} 
\\ \hline
Vanilla & 0.2708 & 0.1487 & 0.2726 & 0.1595 & 0.2349 & 0.1380 & 0.6618 & 8.1808 \\
Prompting (a) & 0.2720 & 0.1492 & 0.2751 & 0.1617 & 0.2376 & 0.1364 & 0.6635 & 8.3344 \\
Prompting (dbrx) & 0.2677 & 0.1449 & 0.2714 & 0.1615 & 0.2333 & 0.1364 & 0.6611 & 7.9563 \\
MemFree Decode & 0.2724 & 0.1468 & 0.2714 & 0.1599 & 0.2383 & 0.1408 & 0.6617 & 8.2453 \\
GA & 0.2423 & 0.1342 & 0.2544 & 0.1502 & 0.2276 & 0.1320 & 0.3102 & 7.5719 \\
NPO & 0.2489 & 0.1367 & 0.2611 & 0.1508 & 0.2335 & 0.1384 & 0.6196 & 8.0313 \\
Gradient Difference & 0.2617 & 0.1425 & 0.2689 & 0.1582 & 0.2374 & 0.1419 & 0.6399 & 4.9438 \\
TV & 0.2394 & 0.1357 & 0.2571 & 0.1507 & 0.2339 & 0.1403 & 0.4887 & 8.1875 \\
SSU & 0.2515 & 0.1364 & 0.2582 & 0.1508 & 0.2409 & 0.1423 & 0.6425 & 8.1415 \\
\hline
\end{tabular}}
\caption{Overall results of Llama3.1 at time step 5, compared with several baselines for $D_f$, $D_{prev}$, and $D_{nor}$. Benchmark performance includes MMLU and MT-Bench scores.}
\vspace{-0.15in}
\label{tab:main_results_time_step_5}
\end{table*}

\begin{table*}[h]
\centering
\resizebox{1\textwidth}{!}{
\begin{tabular}{l||ll||ll||ll||ll}
\hline
\multicolumn{1}{c||}{\multirow{2}{*}{}} & 
\multicolumn{2}{c||}{\textbf{$\mathbf{D_f}$}} & 
\multicolumn{2}{c||}{\textbf{$\mathbf{D_{prev}}$}} & 
\multicolumn{2}{c||}{\textbf{$\mathbf{D_{nor}}$}} &
\multicolumn{2}{c}{\textbf{Benchmark}} \\ \cline{2-9}

\multicolumn{1}{c||}{} &
\begin{tabular}[c]{@{}l@{}}Rouge-1 \\ \end{tabular} &
\begin{tabular}[c]{@{}l@{}}Rouge-L\\ \end{tabular} &
\begin{tabular}[c]{@{}l@{}}Rouge-1\\ \end{tabular} &
\begin{tabular}[c]{@{}l@{}}Rouge-L \\  \end{tabular} &
\begin{tabular}[c]{@{}l@{}}Rouge-1 \\ \end{tabular} &
\begin{tabular}[c]{@{}l@{}}Rouge-L \\  \end{tabular} &
\begin{tabular}[c]{@{}l@{}}MMLU \\ \end{tabular} &
\begin{tabular}[c]{@{}l@{}}MT-Bench \\ \end{tabular} 
\\ \hline
Vanilla & 0.2602 & 0.1472 & 0.2723 & 0.1558 & 0.2349 & 0.1380 & 0.6618 & 8.1808 \\
Prompting (a) & 0.2603 & 0.1478 & 0.2747 & 0.1565 & 0.2376 & 0.1364 & 0.6635 & 8.3344 \\
Prompting (dbrx) & 0.2567 & 0.1489 & 0.2717 & 0.1552 & 0.2333 & 0.1364 & 0.6611 & 7.9563 \\
MemFree Decode & 0.2503 & 0.1452 & 0.2726 & 0.1551 & 0.2383 & 0.1408 & 0.6617 & 8.2453 \\
GA & 0.2535 & 0.1420 & 0.2499 & 0.1430 & 0.2278 & 0.1323 & 0.3082 & 7.6594 \\
NPO & 0.2502 & 0.1419 & 0.2563 & 0.1472 & 0.2342 & 0.1379 & 0.6018 & 8.0375 \\
Gradient Difference & 0.2455 & 0.1391 & 0.2686 & 0.1525 & 0.2369 & 0.1397 & 0.6232 & 4.5500 \\
TV & 0.2408 & 0.1405 & 0.2518 & 0.1459 & 0.2287 & 0.1399 & 0.3116 & 8.1219 \\
SSU & 0.2462 & 0.1398 & 0.2581 & 0.1463 & 0.2374 & 0.1401 & 0.6298 & 8.2359 \\
\hline
\end{tabular}}
\caption{Overall results of Llama3.1 at time step 6, compared with several baselines for $D_f$, $D_{prev}$, and $D_{nor}$. Benchmark performance includes MMLU and MT-Bench scores.}
\vspace{-0.15in}
\label{tab:main_results_time_step_6}
\end{table*}

\begin{table*}[h]
\centering
\resizebox{1\textwidth}{!}{
\begin{tabular}{l||ll||ll||ll||ll}
\hline
\multicolumn{1}{c||}{\multirow{2}{*}{}} & 
\multicolumn{2}{c||}{\textbf{$\mathbf{D_f}$}} & 
\multicolumn{2}{c||}{\textbf{$\mathbf{D_{prev}}$}} & 
\multicolumn{2}{c||}{\textbf{$\mathbf{D_{nor}}$}} &
\multicolumn{2}{c}{\textbf{Benchmark}} \\ \cline{2-9}

\multicolumn{1}{c||}{} &
\begin{tabular}[c]{@{}l@{}}Rouge-1 \\ \end{tabular} &
\begin{tabular}[c]{@{}l@{}}Rouge-L\\ \end{tabular} &
\begin{tabular}[c]{@{}l@{}}Rouge-1\\ \end{tabular} &
\begin{tabular}[c]{@{}l@{}}Rouge-L \\  \end{tabular} &
\begin{tabular}[c]{@{}l@{}}Rouge-1 \\ \end{tabular} &
\begin{tabular}[c]{@{}l@{}}Rouge-L \\  \end{tabular} &
\begin{tabular}[c]{@{}l@{}}MMLU \\ \end{tabular} &
\begin{tabular}[c]{@{}l@{}}MT-Bench \\ \end{tabular} 
\\ \hline
Vanilla & 0.2678 & 0.1482 & 0.2625 & 0.1507 & 0.2349 & 0.1380 & 0.6618 & 8.1808 \\
Prompting (a) & 0.2674 & 0.1521 & 0.2676 & 0.1528 & 0.2376 & 0.1364 & 0.6635 & 8.3344 \\
Prompting (dbrx) & 0.2602 & 0.1489 & 0.2632 & 0.1535 & 0.2333 & 0.1364 & 0.6611 & 7.9563 \\
MemFree Decode & 0.2488 & 0.1451 & 0.2675 & 0.1559 & 0.2383 & 0.1408 & 0.6617 & 8.2453 \\
GA & 0.2485 & 0.1403 & 0.2473 & 0.1441 & 0.2277 & 0.1324 & 0.2729 & 7.7063 \\
NPO & 0.2534 & 0.1452 & 0.2567 & 0.1465 & 0.2369 & 0.1382 & 0.5786 & 8.0375 \\
Gradient Difference & 0.2637 & 0.1494 & 0.2588 & 0.1492 & 0.2386 & 0.1414 & 0.6112 & 4.1384 \\
TV & 0.2453 & 0.1406 & 0.2433 & 0.1419 & 0.2266 & 0.1365 & 0.3477 & 7.8899 \\
SSU & 0.2532 & 0.1428 & 0.2559 & 0.1457 & 0.2391 & 0.1398 & 0.6291 & 8.1406 \\
\hline
\end{tabular}}
\caption{Overall results of Llama3.1 at time step 7, compared with several baselines for $D_f$, $D_{prev}$, and $D_{nor}$. Benchmark performance includes MMLU and MT-Bench scores.}
\vspace{-0.15in}
\label{tab:main_results_time_step_7}
\end{table*}

\begin{table*}[h]
\centering
\resizebox{1\textwidth}{!}{
\begin{tabular}{l||ll||ll||ll||ll}
\hline
\multicolumn{1}{c||}{\multirow{2}{*}{}} & 
\multicolumn{2}{c||}{\textbf{$\mathbf{D_f}$}} & 
\multicolumn{2}{c||}{\textbf{$\mathbf{D_{prev}}$}} & 
\multicolumn{2}{c||}{\textbf{$\mathbf{D_{nor}}$}} &
\multicolumn{2}{c}{\textbf{Benchmark}} \\ \cline{2-9}

\multicolumn{1}{c||}{} &
\begin{tabular}[c]{@{}l@{}}Rouge-1 \\ \end{tabular} &
\begin{tabular}[c]{@{}l@{}}Rouge-L\\ \end{tabular} &
\begin{tabular}[c]{@{}l@{}}Rouge-1\\ \end{tabular} &
\begin{tabular}[c]{@{}l@{}}Rouge-L \\  \end{tabular} &
\begin{tabular}[c]{@{}l@{}}Rouge-1 \\ \end{tabular} &
\begin{tabular}[c]{@{}l@{}}Rouge-L \\  \end{tabular} &
\begin{tabular}[c]{@{}l@{}}MMLU \\ \end{tabular} &
\begin{tabular}[c]{@{}l@{}}MT-Bench \\ \end{tabular} 
\\ \hline
Vanilla & 0.2906 & 0.1673 & 0.2685 & 0.1514 & 0.2349 & 0.1380 & 0.6618 & 8.1808 \\
Prompting (a) & 0.2912 & 0.1668 & 0.2656 & 0.1538 & 0.2376 & 0.1364 & 0.6635 & 8.3344 \\
Prompting (dbrx) & 0.2902 & 0.1648 & 0.2637 & 0.1521 & 0.2333 & 0.1364 & 0.6611 & 7.9563 \\
MemFree Decode & 0.2922 & 0.1690 & 0.2683 & 0.1543 & 0.2383 & 0.1408 & 0.6617 & 8.2453 \\
GA & 0.2623 & 0.1476 & 0.2471 & 0.1451 & 0.2266 & 0.1325 & 0.2674 & 7.7219 \\
NPO & 0.2668 & 0.1516 & 0.2515 & 0.1434 & 0.2368 & 0.1383 & 0.5783 & 8.0719 \\
Gradient Difference & 0.2786 & 0.1578 & 0.2609 & 0.1513 & 0.2393 & 0.1414 & 0.6112 & 4.4000 \\
TV & 0.2539 & 0.1505 & 0.2347 & 0.1388 & 0.2259 & 0.1377 & 0.3516 & 7.9281 \\
SSU & 0.2676 & 0.1493 & 0.2479 & 0.1438 & 0.2386 & 0.1393 & 0.6263 & 8.2344 \\
\hline
\end{tabular}}
\caption{Overall results of Llama3.1 at time step 8, compared with several baselines for $D_f$, $D_{prev}$, and $D_{nor}$. Benchmark performance includes MMLU and MT-Bench scores.}
\vspace{-0.15in}
\label{tab:main_results_time_step_8}
\end{table*}

\begin{table*}[h]
\centering
\resizebox{1\textwidth}{!}{
\begin{tabular}{l||ll||ll||ll||ll}
\hline
\multicolumn{1}{c||}{\multirow{2}{*}{}} & 
\multicolumn{2}{c||}{\textbf{$\mathbf{D_f}$}} & 
\multicolumn{2}{c||}{\textbf{$\mathbf{D_{prev}}$}} & 
\multicolumn{2}{c||}{\textbf{$\mathbf{D_{nor}}$}} &
\multicolumn{2}{c}{\textbf{Benchmark}} \\ \cline{2-9}

\multicolumn{1}{c||}{} &
\begin{tabular}[c]{@{}l@{}}Rouge-1 \\ \end{tabular} &
\begin{tabular}[c]{@{}l@{}}Rouge-L\\ \end{tabular} &
\begin{tabular}[c]{@{}l@{}}Rouge-1\\ \end{tabular} &
\begin{tabular}[c]{@{}l@{}}Rouge-L \\  \end{tabular} &
\begin{tabular}[c]{@{}l@{}}Rouge-1 \\ \end{tabular} &
\begin{tabular}[c]{@{}l@{}}Rouge-L \\  \end{tabular} &
\begin{tabular}[c]{@{}l@{}}MMLU \\ \end{tabular} &
\begin{tabular}[c]{@{}l@{}}MT-Bench \\ \end{tabular} 
\\ \hline
Vanilla & 0.2725 & 0.1628 & 0.2662 & 0.1564 & 0.2349 & 0.1380 & 0.6618 & 8.1808 \\
Prompting (a) & 0.2726 & 0.1601 & 0.2670 & 0.1528 & 0.2376 & 0.1364 & 0.6635 & 8.3344 \\
Prompting (dbrx) & 0.2697 & 0.1579 & 0.2604 & 0.1520 & 0.2333 & 0.1364 & 0.6611 & 7.9563 \\
MemFree Decode & 0.2695 & 0.1583 & 0.2674 & 0.1552 & 0.2383 & 0.1408 & 0.6617 & 8.2453 \\
GA & 0.2608 & 0.1505 & 0.2480 & 0.1441 & 0.2279 & 0.1343 & 0.1996 & 7.4281 \\
NPO & 0.2619 & 0.1530 & 0.2532 & 0.1448 & 0.2299 & 0.1348 & 0.5488 & 8.0281 \\
Gradient Difference & 0.2594 & 0.1531 & 0.2562 & 0.1482 & 0.2422 & 0.1417 & 0.6305 & 4.3208 \\
TV & 0.0001 & 0.0001 & 0.0005 & 0.0005 & 0.0002 & 0.0002 & 0 & 4.2373 \\
SSU & 0.2629 & 0.1522 & 0.2484 & 0.1427 & 0.2360 & 0.1382 & 0.6049 & 8.1219 \\
\hline
\end{tabular}}
\caption{Overall results of Llama3.1 at time step 9, compared with several baselines for $D_f$, $D_{prev}$, and $D_{nor}$. Benchmark performance includes MMLU and MT-Bench scores.}
\vspace{-0.15in}
\label{tab:main_results_time_step_9}
\end{table*}

\begin{table*}[h]
\centering
\resizebox{1\textwidth}{!}{
\begin{tabular}{l||ll||ll||ll||ll}
\hline
\multicolumn{1}{c||}{\multirow{2}{*}{}} & 
\multicolumn{2}{c||}{\textbf{$\mathbf{D_f}$}} & 
\multicolumn{2}{c||}{\textbf{$\mathbf{D_{prev}}$}} & 
\multicolumn{2}{c||}{\textbf{$\mathbf{D_{nor}}$}} &
\multicolumn{2}{c}{\textbf{Benchmark}} \\ \cline{2-9}

\multicolumn{1}{c||}{} &
\begin{tabular}[c]{@{}l@{}}Rouge-1 \\ \end{tabular} &
\begin{tabular}[c]{@{}l@{}}Rouge-L\\ \end{tabular} &
\begin{tabular}[c]{@{}l@{}}Rouge-1\\ \end{tabular} &
\begin{tabular}[c]{@{}l@{}}Rouge-L \\  \end{tabular} &
\begin{tabular}[c]{@{}l@{}}Rouge-1 \\ \end{tabular} &
\begin{tabular}[c]{@{}l@{}}Rouge-L \\  \end{tabular} &
\begin{tabular}[c]{@{}l@{}}MMLU \\ \end{tabular} &
\begin{tabular}[c]{@{}l@{}}MT-Bench \\ \end{tabular} 
\\ \hline
Vanilla & 0.2667 & 0.1458 & 0.2602 & 0.1467 & 0.2349 & 0.1380 & 0.6618 & 8.1808 \\
Prompting (a) & 0.2605 & 0.1469 & 0.2597 & 0.1477 & 0.2376 & 0.1364 & 0.6635 & 8.3344 \\
Prompting (dbrx) & 0.2622 & 0.1467 & 0.2648 & 0.1510 & 0.2333 & 0.1364 & 0.6611 & 7.9563 \\
MemFree Decode & 0.2596 & 0.1450 & 0.2672 & 0.1522 & 0.2383 & 0.1408 & 0.6617 & 8.2453 \\
GA & 0.2559 & 0.1422 & 0.2491 & 0.1422 & 0.2289 & 0.1335 & 0.2004 & 7.4344 \\
NPO & 0.2583 & 0.1435 & 0.2602 & 0.1498 & 0.2306 & 0.1342 & 0.5477 & 7.9969 \\
Gradient Difference & 0.2542 & 0.1453 & 0.2601 & 0.1496 & 0.2365 & 0.1383 & 0.6079 & 4.4843 \\
TV & 0 & 0 & 0.0005 & 0.0005 & 0.0002 & 0.0002 & 0 & 3.915 \\
SSU & 0.2481 & 0.1389 & 0.2541 & 0.1439 & 0.2333 & 0.1396 & 0.6023 & 8.0206 \\
\hline
\end{tabular}}
\caption{Overall results of Llama3.1 at time step 10, compared with several baselines for $D_f$, $D_{prev}$, and $D_{nor}$. Benchmark performance includes MMLU and MT-Bench scores.}
\vspace{-0.15in}
\label{tab:main_results_time_step_10}
\end{table*}

\begin{table*}[h]
\centering
\resizebox{1\textwidth}{!}{
\begin{tabular}{l||ll||ll||ll||ll}
\hline
\multicolumn{1}{c||}{\multirow{2}{*}{}} & 
\multicolumn{2}{c||}{\textbf{$\mathbf{D_f}$}} & 
\multicolumn{2}{c||}{\textbf{$\mathbf{D_{prev}}$}} & 
\multicolumn{2}{c||}{\textbf{$\mathbf{D_{nor}}$}} &
\multicolumn{2}{c}{\textbf{Benchmark}} \\ \cline{2-9}

\multicolumn{1}{c||}{} &
\begin{tabular}[c]{@{}l@{}}Rouge-1 \\ \end{tabular} &
\begin{tabular}[c]{@{}l@{}}Rouge-L\\ \end{tabular} &
\begin{tabular}[c]{@{}l@{}}Rouge-1\\ \end{tabular} &
\begin{tabular}[c]{@{}l@{}}Rouge-L \\  \end{tabular} &
\begin{tabular}[c]{@{}l@{}}Rouge-1 \\ \end{tabular} &
\begin{tabular}[c]{@{}l@{}}Rouge-L \\  \end{tabular} &
\begin{tabular}[c]{@{}l@{}}MMLU \\ \end{tabular} &
\begin{tabular}[c]{@{}l@{}}MT-Bench \\ \end{tabular} 
\\ \hline
Vanilla & 0.2828 & 0.1629 & 0 & 0 & 0.2456 & 0.1487 & 0.6070 & 7.4563 \\
Prompting (a) & 0.2731 & 0.1596 & 0 & 0 & 0.2481 & 0.1515 & 0.6074 & 7.1438 \\
Prompting (dbrx) & 0.2783 & 0.1649 & 0 & 0 & 0.2412 & 0.1482 & 0.6053 & 7.3531 \\
MemFree Decode & 0.2742 & 0.1640 & 0 & 0 & 0.2458 & 0.1502 & 0.6074 & 7.1719 \\
GA & 0.0739 & 0.0501 & 0 & 0 & 0.1183 & 0.0750 & 0.6028 & 6.5875 \\
NPO & 0.2721 & 0.1561 & 0 & 0 & 0.2447 & 0.1466 & 0.6028 & 7.5547 \\
Gradient Difference & 0.2472 & 0.1462 & 0 & 0 & 0.2351 & 0.1413 & 0.6074 & 7.3875 \\
TV & 0.2591 & 0.1539 & 0 & 0 & 0.2391 & 0.1469 & 0.6021 & 7.1125 \\
SSU & 0.2536 & 0.1536 & 0 & 0 & 0.2401 & 0.1495 & 0.6052 & 7.4406 \\
\hline
\end{tabular}}
\caption{Overall results of Mistral-7B at time step 1, compared with several baselines for $D_f$, $D_{prev}$, and $D_{nor}$. Benchmark performance includes MMLU and MT-Bench scores.}
\vspace{-0.15in}
\label{tab:main_results_mistral_time_step_1}
\end{table*}

\begin{table*}[h]
\centering
\resizebox{1\textwidth}{!}{
\begin{tabular}{l||ll||ll||ll||ll}
\hline
\multicolumn{1}{c||}{\multirow{2}{*}{}} & 
\multicolumn{2}{c||}{\textbf{$\mathbf{D_f}$}} & 
\multicolumn{2}{c||}{\textbf{$\mathbf{D_{prev}}$}} & 
\multicolumn{2}{c||}{\textbf{$\mathbf{D_{nor}}$}} &
\multicolumn{2}{c}{\textbf{Benchmark}} \\ \cline{2-9}

\multicolumn{1}{c||}{} &
\begin{tabular}[c]{@{}l@{}}Rouge-1 \\ \end{tabular} &
\begin{tabular}[c]{@{}l@{}}Rouge-L\\ \end{tabular} &
\begin{tabular}[c]{@{}l@{}}Rouge-1\\ \end{tabular} &
\begin{tabular}[c]{@{}l@{}}Rouge-L \\  \end{tabular} &
\begin{tabular}[c]{@{}l@{}}Rouge-1 \\ \end{tabular} &
\begin{tabular}[c]{@{}l@{}}Rouge-L \\  \end{tabular} &
\begin{tabular}[c]{@{}l@{}}MMLU \\ \end{tabular} &
\begin{tabular}[c]{@{}l@{}}MT-Bench \\ \end{tabular} 
\\ \hline
Vanilla & 0.2997 & 0.1838 & 0.2797 & 0.1650 & 0.2456 & 0.1487 & 0.6070 & 7.4563 \\
Prompting (a) & 0.3038 & 0.1842 & 0.2776 & 0.1653 & 0.2481 & 0.1515 & 0.6074 & 7.1438 \\
Prompting (dbrx) & 0.3087 & 0.1867 & 0.2850 & 0.1672 & 0.2412 & 0.1482 & 0.6053 & 7.3531 \\
MemFree Decode & 0.3021 & 0.1841 & 0.2809 & 0.1661 & 0.2458 & 0.1502 & 0.6074 & 7.1719 \\
GA & 0.0544 & 0.0357 & 0.0177 & 0.0118 & 0.0347 & 0.0233 & 0.6039 & 4.0313 \\
NPO & 0.2977 & 0.1779 & 0.2764 & 0.1605 & 0.2453 & 0.1474 & 0.6028 & 7.3438 \\
Gradient Difference & 0.2908 & 0.1728 & 0.2668 & 0.1505 & 0.2373 & 0.1412 & 0.6077 & 7.5313 \\
TV & 0.2682 & 0.1672 & 0.2454 & 0.1451 & 0.2258 & 0.1377 & 0.5944 & 7.0469 \\
SSU & 0.2871 & 0.1726 & 0.2600 & 0.1537 & 0.2357 & 0.1451 & 0.6028 & 7.3000 \\
\hline
\end{tabular}}
\caption{Overall results of Mistral-7B at time step 2, compared with several baselines for $D_f$, $D_{prev}$, and $D_{nor}$. Benchmark performance includes MMLU and MT-Bench scores.}
\vspace{-0.15in}
\label{tab:main_results_mistral_time_step_2}
\end{table*}

\begin{table*}[h]
\centering
\resizebox{1\textwidth}{!}{
\begin{tabular}{l||ll||ll||ll||ll}
\hline
\multicolumn{1}{c||}{\multirow{2}{*}{}} & 
\multicolumn{2}{c||}{\textbf{$\mathbf{D_f}$}} & 
\multicolumn{2}{c||}{\textbf{$\mathbf{D_{prev}}$}} & 
\multicolumn{2}{c||}{\textbf{$\mathbf{D_{nor}}$}} &
\multicolumn{2}{c}{\textbf{Benchmark}} \\ \cline{2-9}

\multicolumn{1}{c||}{} &
\begin{tabular}[c]{@{}l@{}}Rouge-1 \\ \end{tabular} &
\begin{tabular}[c]{@{}l@{}}Rouge-L\\ \end{tabular} &
\begin{tabular}[c]{@{}l@{}}Rouge-1\\ \end{tabular} &
\begin{tabular}[c]{@{}l@{}}Rouge-L \\  \end{tabular} &
\begin{tabular}[c]{@{}l@{}}Rouge-1 \\ \end{tabular} &
\begin{tabular}[c]{@{}l@{}}Rouge-L \\  \end{tabular} &
\begin{tabular}[c]{@{}l@{}}MMLU \\ \end{tabular} &
\begin{tabular}[c]{@{}l@{}}MT-Bench \\ \end{tabular} 
\\ \hline
Vanilla & 0.2523 & 0.1562 & 0.2943 & 0.1755 & 0.2456 & 0.1487 & 0.6070 & 7.4563 \\
Prompting (a) & 0.2440 & 0.1518 & 0.2928 & 0.1742 & 0.2481 & 0.1515 & 0.6074 & 7.1438 \\
Prompting (dbrx) & 0.2429 & 0.1561 & 0.2883 & 0.1746 & 0.2412 & 0.1482 & 0.6053 & 7.3531 \\
MemFree Decode & 0.2401 & 0.1527 & 0.2906 & 0.1758 & 0.2458 & 0.1502 & 0.6074 & 7.1719 \\
GA & 0.0313 & 0.0205 & 0.0274 & 0.0193 & 0.0323 & 0.0216 & 0.6053 & 3.9056 \\
NPO & 0.2448 & 0.1486 & 0.2847 & 0.1677 & 0.2434 & 0.1474 & 0.6046 & 7.2438 \\
Gradient Difference & 0.2392 & 0.1405 & 0.2719 & 0.1505 & 0.2327 & 0.1383 & 0.6049 & 7.2844 \\
TV & 0.2054 & 0.1319 & 0.2404 & 0.1479 & 0.2095 & 0.1314 & 0.5846 & 6.8469 \\
SSU & 0.2204 & 0.1390 & 0.2623 & 0.1601 & 0.2333 & 0.1426 & 0.5996 & 7.3437 \\
\hline
\end{tabular}}
\caption{Overall results of Mistral-7B at time step 3, compared with several baselines for $D_f$, $D_{prev}$, and $D_{nor}$. Benchmark performance includes MMLU and MT-Bench scores.}
\vspace{-0.15in}
\label{tab:main_results_mistral_time_step_3}
\end{table*}

\begin{table*}[h]
\centering
\resizebox{1\textwidth}{!}{
\begin{tabular}{l||ll||ll||ll||ll}
\hline
\multicolumn{1}{c||}{\multirow{2}{*}{}} & 
\multicolumn{2}{c||}{\textbf{$\mathbf{D_f}$}} & 
\multicolumn{2}{c||}{\textbf{$\mathbf{D_{prev}}$}} & 
\multicolumn{2}{c||}{\textbf{$\mathbf{D_{nor}}$}} &
\multicolumn{2}{c}{\textbf{Benchmark}} \\ \cline{2-9}

\multicolumn{1}{c||}{} &
\begin{tabular}[c]{@{}l@{}}Rouge-1 \\ \end{tabular} &
\begin{tabular}[c]{@{}l@{}}Rouge-L\\ \end{tabular} &
\begin{tabular}[c]{@{}l@{}}Rouge-1\\ \end{tabular} &
\begin{tabular}[c]{@{}l@{}}Rouge-L \\  \end{tabular} &
\begin{tabular}[c]{@{}l@{}}Rouge-1 \\ \end{tabular} &
\begin{tabular}[c]{@{}l@{}}Rouge-L \\  \end{tabular} &
\begin{tabular}[c]{@{}l@{}}MMLU \\ \end{tabular} &
\begin{tabular}[c]{@{}l@{}}MT-Bench \\ \end{tabular} 
\\ \hline
Vanilla & 0.2702 & 0.1660 & 0.2929 & 0.1753 & 0.2456 & 0.1487 & 0.6070 & 7.4563 \\
Prompting (a) & 0.2730 & 0.1650 & 0.2863 & 0.1739 & 0.2481 & 0.1515 & 0.6074 & 7.1438 \\
Prompting (dbrx) & 0.2726 & 0.1687 & 0.2817 & 0.1726 & 0.2412 & 0.1482 & 0.6053 & 7.3531 \\
MemFree Decode & 0.2711 & 0.1628 & 0.2846 & 0.1748 & 0.2458 & 0.1502 & 0.6074 & 7.1719 \\
GA & 0 & 0 & 0 & 0 & 0 & 0 & 0 & 0 \\
NPO & 0.2618 & 0.1561 & 0.2806 & 0.1665 & 0.2447 & 0.1475 & 0.6021 & 7.2688 \\
Gradient Difference & 0.2503 & 0.1464 & 0.2694 & 0.1595 & 0.2326 & 0.1408 & 0.6042 & 7.2375 \\
TV & 0.2038 & 0.1266 & 0.2131 & 0.1306 & 0.1929 & 0.1189 & 0.5582 & 6.4938 \\
SSU & 0.2431 & 0.1489 & 0.2553 & 0.1561 & 0.2352 & 0.1434 & 0.6000 & 7.2344 \\
\hline
\end{tabular}}
\caption{Overall results of Mistral-7B at time step 4, compared with several baselines for $D_f$, $D_{prev}$, and $D_{nor}$. Benchmark performance includes MMLU and MT-Bench scores.}
\vspace{-0.15in}
\label{tab:main_results_mistral_time_step_4}
\end{table*}

\begin{table*}[h]
\centering
\resizebox{1\textwidth}{!}{
\begin{tabular}{l||ll||ll||ll||ll}
\hline
\multicolumn{1}{c||}{\multirow{2}{*}{}} & 
\multicolumn{2}{c||}{\textbf{$\mathbf{D_f}$}} & 
\multicolumn{2}{c||}{\textbf{$\mathbf{D_{prev}}$}} & 
\multicolumn{2}{c||}{\textbf{$\mathbf{D_{nor}}$}} &
\multicolumn{2}{c}{\textbf{Benchmark}} \\ \cline{2-9}

\multicolumn{1}{c||}{} &
\begin{tabular}[c]{@{}l@{}}Rouge-1 \\ \end{tabular} &
\begin{tabular}[c]{@{}l@{}}Rouge-L\\ \end{tabular} &
\begin{tabular}[c]{@{}l@{}}Rouge-1\\ \end{tabular} &
\begin{tabular}[c]{@{}l@{}}Rouge-L \\  \end{tabular} &
\begin{tabular}[c]{@{}l@{}}Rouge-1 \\ \end{tabular} &
\begin{tabular}[c]{@{}l@{}}Rouge-L \\  \end{tabular} &
\begin{tabular}[c]{@{}l@{}}MMLU \\ \end{tabular} &
\begin{tabular}[c]{@{}l@{}}MT-Bench \\ \end{tabular} 
\\ \hline
Vanilla & 0.2709 & 0.1555 & 0.2829 & 0.1703 & 0.2456 & 0.1487 & 0.6070 & 7.4563 \\
Prompting (a) & 0.2673 & 0.1530 & 0.2904 & 0.1719 & 0.2481 & 0.1515 & 0.6074 & 7.1438 \\
Prompting (dbrx) & 0.2673 & 0.1561 & 0.2835 & 0.1714 & 0.2412 & 0.1482 & 0.6053 & 7.3531 \\
MemFree Decode & 0.2624 & 0.1515 & 0.2833 & 0.1735 & 0.2458 & 0.1502 & 0.6074 & 7.1719 \\
GA & 0 & 0 & 0 & 0 & 0 & 0 & 0 & 0 \\
NPO & 0.2633 & 0.1490 & 0.2819 & 0.1647 & 0.2423 & 0.1462 & 0.5986 & 7.4719 \\
Gradient Difference & 0.2503 & 0.1464 & 0.0483 & 0.0336 & 0.0491 & 0.0338 & 0.6063 & 2.8375 \\
TV & 0.1703 & 0.1078 & 0.1674 & 0.1099 & 0.1468 & 0.0986 & 0 & 1.0000 \\
SSU & 0.2324 & 0.1402 & 0.2571 & 0.1575 & 0.2333 & 0.1427 & 0.6000 & 7.3469 \\
\hline
\end{tabular}}
\caption{Overall results of Mistral-7B at time step 5, compared with several baselines for $D_f$, $D_{prev}$, and $D_{nor}$. Benchmark performance includes MMLU and MT-Bench scores.}
\vspace{-0.15in}
\label{tab:main_results_mistral_time_step_5}
\end{table*}

\begin{table*}[h]
\centering
\resizebox{1\textwidth}{!}{
\begin{tabular}{l||ll||ll||ll||ll}
\hline
\multicolumn{1}{c||}{\multirow{2}{*}{}} & 
\multicolumn{2}{c||}{\textbf{$\mathbf{D_f}$}} & 
\multicolumn{2}{c||}{\textbf{$\mathbf{D_{prev}}$}} & 
\multicolumn{2}{c||}{\textbf{$\mathbf{D_{nor}}$}} &
\multicolumn{2}{c}{\textbf{Benchmark}} \\ \cline{2-9}

\multicolumn{1}{c||}{} &
\begin{tabular}[c]{@{}l@{}}Rouge-1 \\ \end{tabular} &
\begin{tabular}[c]{@{}l@{}}Rouge-L\\ \end{tabular} &
\begin{tabular}[c]{@{}l@{}}Rouge-1\\ \end{tabular} &
\begin{tabular}[c]{@{}l@{}}Rouge-L \\  \end{tabular} &
\begin{tabular}[c]{@{}l@{}}Rouge-1 \\ \end{tabular} &
\begin{tabular}[c]{@{}l@{}}Rouge-L \\  \end{tabular} &
\begin{tabular}[c]{@{}l@{}}MMLU \\ \end{tabular} &
\begin{tabular}[c]{@{}l@{}}MT-Bench \\ \end{tabular} 
\\ \hline
Vanilla & 0.2694 & 0.1610 & 0.2778 & 0.1628 & 0.2456 & 0.1487 & 0.6070 & 7.4563 \\
Prompting (a) & 0.2653 & 0.1580 & 0.2806 & 0.1632 & 0.2481 & 0.1515 & 0.6074 & 7.1438 \\
Prompting (dbrx) & 0.2692 & 0.1617 & 0.2862 & 0.1679 & 0.2412 & 0.1482 & 0.6053 & 7.3531 \\
MemFree Decode & 0.2662 & 0.1585 & 0.2758 & 0.1638 & 0.2458 & 0.1502 & 0.6074 & 7.1719 \\
GA & 0 & 0 & 0 & 0 & 0 & 0 & 0 & 0 \\
NPO & 0.2711 & 0.1572 & 0.2714 & 0.1555 & 0.2457 & 0.1482 & 0.5979 & 7.4119 \\
Gradient Difference & 0.0273 & 0.0191 & 0.0270 & 0.0198 & 0.0485 & 0.0326 & 0.6070 & 2.6815 \\
TV & 0 & 0 & 0 & 0 & 0 & 0 & 0 & 0 \\
SSU & 0.2519 & 0.1476 & 0.2448 & 0.1463 & 0.2304 & 0.1416 & 0.5982 & 7.2938 \\
\hline
\end{tabular}}
\caption{Overall results of Mistral-7B at time step 6, compared with several baselines for $D_f$, $D_{prev}$, and $D_{nor}$. Benchmark performance includes MMLU and MT-Bench scores.}
\vspace{-0.15in}
\label{tab:main_results_mistral_time_step_6}
\end{table*}

\chapter{Limitations and Future Directions}
\section{Lack of Robust Evaluation for Machine Unlearning}
In this thesis, we primarily use lexical-based evaluation metrics to assess the algorithm's performance. However, as \citet{ippolito2023preventing} notes, measuring verbatim memorization can create a false sense of privacy. Furthermore,  \citet{lucki2024adversarial} and  \citet{patil2023can} highlight that existing unlearning and editing methods may merely obfuscate data rather than achieve genuine unlearning. 

To rigorously evaluate unlearning, it is essential to go beyond Rouge-1 and Rouge-L scores and incorporate methods that can effectively detect potential training data leakage. Membership Inference Attacks (MIAs) \citep{shokri2017membership} offer a promising avenue for this purpose. However, recent studies suggest that the performance of MIAs is close to random guessing for pre-trained LLMs under various conditions \citep{duan2024membership, yao2024machine}. 

Additionally, developing a unified jailbreak-type attack using unlearned knowledge could provide a robust means to evaluate the effectiveness of unlearning algorithms. This approach is crucial for advancing research in this field. We encourage future work to design more effective MIAs and apply them to the sequential unlearning setting proposed in this study.

\section{Challenges in Knowledge and Capability Retention}
Although \method achieves a better trade-off between unlearning efficacy and general-purpose language abilities retention compared to other baseline methods, we still observe some unintended knowledge loss in books that are not targeted for unlearning, as well as a reduction in the model's reasoning and conversation abilities. Future work should focus on further minimizing the gap in knowledge and capability retention between the modified model and the original, in order to ensure better general knowledge preservation during sequential unlearning.

\section{The Need for Certified Unlearning Mechanisms}
Our thesis does not provide a theoretical guarantee to certify unlearning. Recently, \citet{zhang2024towards} explored certified unlearning for deep neural networks in non-convex settings by establishing a connection between unlearning and differential privacy~\citep{dwork2006differential}. A similar certified mechanism is also necessary for generative models, and we leave this as an avenue for future work.

\section{Unlearning is Not All You Need} 
Reducing the risks of copyright infringement involves more than just unlearning methods. For instance, \citet{liu2024infini} introduces Infini-gram, an efficient search engine designed to scan the massive training corpora of LLMs. An alternative approach could involve using tools like Infini-gram to identify copyrighted passages that are being generated and then intercede to have the model regenerate non-copyrighted output. This kind generation-time copyright takedown method presents another promising direction for exploration. Developing lookahead techniques to detect whether copyrighted content is likely to be generated, and implementing alternative generation strategies to prevent such outputs, could be a interesting direction for future research.

Future research could focus on expanding such search engine tools to cover a broader range of datasets used in training state-of-the-art LLMs. Additionally, regularly updating model checkpoints to comply with \textit{the right to be forgotten} would be an essential step toward ensuring ongoing compliance with legal and ethical standards.

Lastly, while unlearning can effectively remove harmful knowledge, private information, and copyrighted content, current unlearning methods often result in reduced general-purpose language abilities, making them less practical for direct use. One hypothesis worth exploring is reapplying instruction tuning after implementing unlearning algorithms. This approach, particularly when applied in a sample-efficient manner\citep{hewitt2024instruction}, could enhance the performance of unlearned LLMs.

More broadly, unlearning methods should be seen as one component of a comprehensive effort to develop safe and trustworthy AI systems \citep{chen2024trustworthy}. Future research could focus on advancing safety, privacy, and alignment in generative models, including model merging \citep{yang2024model}, defenses against various attacks, watermarking techniques, and reinforcement learning methods to improve alignment with human values.

\chapter{Summary}
In this thesis, we investigate the sequential unlearning of copyrighted content from LLMs as a means to mitigate copyright infringement. We propose \method, a novel approach that leverages random labeling loss and gradient-based weight saliency to enable more stable sequential unlearning. Through extensive experiments, we demonstrate that \method achieves a superior balance between unlearning efficacy and the retention of general-purpose language abilities compared to existing unlearning techniques and other copyright takedown methods. Additionally, ablation studies validate the contributions of individual components of \method.

Finally, we address several limitations and propose directions for future research. These include refining evaluation metrics for unlearning methods, optimizing the trade-off between unlearning efficacy and the retention of general-purpose language abilities, establishing theoretical guarantees by linking unlearning techniques to Differential Privacy, and investigating alternative approaches to mitigating copyright infringement risks in generative models.

\end{mainf}
%-------------------------------------------

%%%%% OPTIONAL APPENDICES %%%%%
%-------------------------------------------
% \begin{append}
% \chapter{\MakeUppercase{Title of Appendix A}}
% The content of Appendix A begins here. Use the \verb|\chapter| command to insert additional appendices.
% \section{Section Name of Appendix A}
% Use the \verb|\section| command to create sections.
% \chapter{\MakeUppercase{Title of Appendix B}}
% The content of Appendix B begins here.
% \end{append}
%-------------------------------------------

%%%%% BIBLIOGRAPHY %%%%%
%-------------------------------------------
\begin{bibliof}
%\nocite{*} % If applicable, uncomment this line to display all entries in the .bib file
\bibliography{bibliography}
\end{bibliof}
%-------------------------------------------
\end{document}